\documentclass[10pt,twocolumn,letterpaper]{article}

\usepackage[pagenumbers]{cvpr}      %

\usepackage{cuted} %

\usepackage{currfile} %

\usepackage{caption} %

\usepackage{tabularx}
\usepackage{algorithm}
\usepackage{algorithmic}
\usepackage{newfloat}
\usepackage{listings}
\DeclareCaptionStyle{ruled}{labelfont=normalfont,labelsep=colon,strut=off}
\floatstyle{ruled}
\newfloat{listing}{tb}{lst}{}
\floatname{listing}{Listing}

\newcommand{\method}{\textsc{ISRS-DETR}\xspace}

\definecolor{cvprblue}{rgb}{0.21,0.49,0.74}
\usepackage[pagebackref,breaklinks,colorlinks,allcolors=cvprblue]{hyperref}

\usepackage{booktabs} \usepackage{multirow}

\def\confName{CVPR}
\def\confYear{2027}

\title{ISRS-DETR: Detection-Guided Click Propagation for Remote Sensing Interactive Segmentation}

\author{
Thanh Duc Pham$^{1}$ \quad
Anh Nguyen$^{2}$ \quad
Duong Duc Hieu$^{1}$ \quad
Minh-Tan Pham$^{3}$\thanks{Corresponding author.}\\
$^{1}$FPT Software AI Center, Vietnam\\
$^{2}$Department of Computer Science, University of Liverpool, UK\\
$^{3}$IRISA, Universit\'e Bretagne Sud, UMR 6074, 56000 Vannes, France\\
{\tt\small thanhpd29@fpt.com, minh-tan.pham@irisa.fr}
}

\begin{document}

\maketitle

\begin{abstract}
Interactive segmentation reduces the prohibitive cost of pixel-level annotation by allowing users to delineate objects with a few clicks. However, applying this paradigm directly to remote sensing imagery is non-trivial: ultra-high resolutions, small object sizes, and sparse spatial distributions all degrade segmentation quality. Recent work has addressed the resolution barrier and achieved competitive results in interactive segmentation for remote sensing (ISRS). However, they treat all instances of a class within an image as a single objective target. Consequently, interactions spent on one object contribute nothing to its same-class neighbours, and satisfactory masks may hinder the practicality of these frameworks. We observe that remote sensing scenes exhibit markedly strong inter-object correlation, meaning a single clicked object is highly informative about the rest of its category. Building on this, we propose \method, a detection-guided interactive segmentation framework that injects object-level evidence into both training and inference. Our \method employs a looped DETR-based decoder, a single weight-shared decoder layer iterated several times, with the interactive segmentation backbone to localise co-occurring same-class objects, and introduces a Dynamic Top-K Click Selection strategy that retains only reliable proposals and converts each into a simulated click, so one user interaction propagates across an entire class. Experiments on three standard remote sensing benchmarks show that \method achieves state-of-the-art accuracy while substantially reducing Number of Clicks (NoC). Code will be released for reproducibility upon acceptance.

\end{abstract}

\section{Introduction}

Segmentation is a common computer vision task in the real world implementations, and in many applications such as medical, these demands emerge simultaneously. In spite of great success in varied visual data, traditional image segmentation commonly require substantial precise annotated data. Nevertheless, acquiring pixel-level annotations is not only labor-intensive but also time-consuming. To this end, interactive segmentation is proposed to address this problem \cite{xu2016deep}. It enables users to select objects and delineate them easily with minimal user interaction. Numerous interactive image segmentation approaches have been widely applied to annotate large-scale image datasets, which support the success of deep models in various applications, including autonomous driving \cite{cordts2016cityscapesdatasetsemanticurban}, and medical imaging \cite{isensee2021nnunet}.
\begin{figure}[t]
  \centering
  \includegraphics[width=\linewidth]{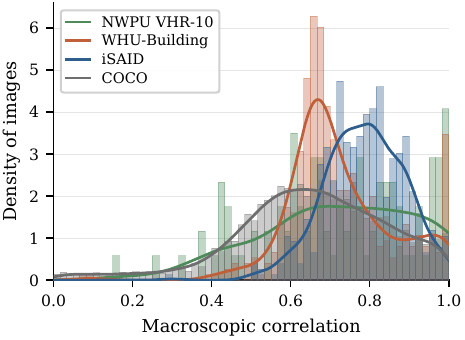}
  \caption{Statistical distribution of macroscopic correlation (MC) on the three
  remote sensing benchmarks compared with the
  MS-COCO benchmark. The remote sensing datasets
  concentrate at markedly higher MC values, indicating stronger inter-object
  position dependence than natural-image scenes.}
  \label{fig:mc_distribution}
\end{figure}

 However, applying interactive segmentation directly to remote sensing remains challenging. Compared with natural images, remote sensing imagery poses unique challenges for interactive segmentation. In particular, ultra-high resolution, small objects and sparse layouts are recognised as a key factor contributing to the performance degradation \cite{cheng2017remote,zhu2017deep, audebert2018beyond} observed in segmentation models. Due to the importance and the difficulty, \textbf{interactive segmentation} in \textbf{remote sensing image} (\textbf{ISRS}) has attracted much research attention. Recently, \citet{lin2026crosscut} introduced CrossCut, an interactive framework that divides the image into patches and encodes the click map into a global prompt, which is subsequently injected back into each patch to facilitate information exchange.

CrossCut is the first to handle ultra-high resolution effectively in ISRS, but it has two limitations. (i) First, Crosscut overlooks a fundamental characteristic of remote sensing imagery: the relationship between multiple objects within the same image. In particular, during both training and inference, they considered all samples of the same class as a single segmentation target. As a result, user interactions performed on one object provide no benefit for segmenting other instances of the same category, despite the strong semantic and visual relationships among them. (ii) Second, despite achieving competitive performance in terms of interaction efficiency, CrossCut still requires numerous clicks to obtain satisfactory segmentation masks. This substantial interaction cost limits the efficiency and scalability of existing ISRS systems in practical implementations.

Moreover, following \citet{hou2024relationdetr}, we measure the macroscopic correlation (MC) between the representations of objects within the same image among four datasets (detailed in Appendix~\ref{app:mc}). As can be shown in Fig.~\ref{fig:mc_distribution}, remote sensing scenes exhibit markedly strong inter-object correlation, indicating that a clicked object is highly informative about its same-class neighbours.

Motivated by the above observations, we introduce \method, a detection-guided interactive segmentation model built on a DETR-based detector, with dedicated components to address in ISRS setting. Unlike conventional approaches that treats all instances of the same class as a single segmentation objective, \method further incorporates object-level representations during both training and inference, enabling interactions on one object to benefit the segmentation of other semantically related instances.

First, we integrate a looped version of  decoder of the RF-DETR-XL \cite{robinson2026rfdetr} with the interactive segmentation framework in order to localize the bounding boxes of same-class objects co-occurring within the image. Second, inspired by \cite{yang2024loopedtransformersbetterlearning, deqdet}, instead of stacking several independently parameterized decoder layers, we iteratively apply a single weight-shared decoder layer $T$ times. Since each DETR decoder layer performs the same iterative box-refinement operation \cite{deformable, dino, peng2024dfineredefineregressiontask}, our looped design not only preserves the detection branch's iterative refinement capability but also introduces only a few of parameters of a single decoder layer into the interactive segmentation model. Third, each detected same-class box is converted into candidate click locations, and our Dynamic Top-K Simulated Click Selection keeps only the candidates whose features are reliably similar to the user's click, with the number kept chosen per image rather than fixed.

Extensive experiments on three challenging remote sensing interactive segmentation benchmarks demonstrate that \method consistently outperforms existing state-of-the-art methods. All quantitative comparisons and qualitative visualizations verify the effectiveness of our object-aware design in improving segmentation quality while substantially reducing user interactions. These results establish \method as a new state-of-the-art framework for interactive segmentation in remote sensing imagery.

Overall, our contributions are summarized as follows:
\begin{itemize}

    \item We identify and empirically characterize a fundamental limitation of existing ISRS methods: they collapse all same-class instances into a single segmentation target, thereby overlooking inter-object relationships. To address this limitation, we propose \method, a detection-guided interactive segmentation framework for remote sensing imagery.

    \item  To the best of our knowledge, we are the first to replace the stacked decoder of a deformable-attention real-time DETR with a single weight-shared layer applied iteratively. This design preserves iterative box refinement while substantially reducing the additional parameter overhead.

    \item We introduce Dynamic Top-K Simulated Click Selection, which transfers the user's click to every detected same-class instance under four orientation hypotheses and keeps an image-adaptive number of the most appropriate candidates.

    \item Extensive experiments on three remote sensing
    benchmarks show that \method achieves state-of-the-art segmentation quality while
    substantially reducing the required user interactions.
\end{itemize}

\section{Related Work}
\subsection{Interactive Image Segmentation.}

Recently, interactive image segmentation has witnessed remarkable advancements. Early frameworks \cite{boykov2006graphcuts,
blake2004interactive, rother2004grabcut, vicente2008graph,
veksler2008star} addressed this challenge as a graph-based optimization problem. However, these traditional methods primarily rely on handcrafted features, resulting in suboptimal performance. With the rise of deep learning, \citet{xu2016deep} first introduced a click simulation strategy and preprocessed the simulated clicks via a distance transform and combining them with the original image as the model input. Many recent work have attempted to integrate ViTs into interactive segmentation \cite{chen2022focalclick,liu2023simpleclick,lee2024mfp}, achieving competitive performance on natural images. Another line of works is SAM \cite{kirillov2023segment_sam}, which was trained on a large dataset and supports flexible prompts such as points and boxes. These advances highlight the increasing attention devoted to interactive segmentation.

\subsection{Remote Sensing Image Segmentation}

Remote sensing image segmentation plays a fundamental role in earth observation and geospatial science. Numerous research has been devoted to address this challenge. Conventional CNN-based methods have achieved remarkable success by leveraging large-scale pixel-wise annotations. However, acquiring such annotations is labor-intensive, time-consuming, and expensive, particularly for high-resolution remote sensing images that contain numerous small objects. Prior work has been done to reduce the annotation cost of labeling images in remote sensing imagery such as: semi-supervised and interactive segmentation. \cite{shan2025rossamhighqualityinteractivesegmentation} adapts SAM \cite{kirillov2023segment_sam} to remote sensing imagery through LoRA fine-tuning and boundary refinement, improving mask quality for moving objects. Moreover, AerOSeg \cite{dutta2025aerosegharnessingsamopenvocabulary} enhances open-vocabulary remote sensing segmentation by integrating SAM guidance, orientation-invariant CLIP features, and semantic-preserving refinement. CrossCut \cite{lin2026crosscut} using flexible patch division strategies to creating global click prompt, that effectively captures global information across patches. Despite all the progress, existing methods largely overlook the relationships among objects that co-occur within the same image.

\subsection{DETR-based Detectors}

\citet{detr} first proposed an end-to-end Transformer-based object detector (DETR), which eliminates the need for hand-crafted anchors and non-maximum suppression (NMS). Although DETR has achieved competitive performance, it suffers from slow convergence, high computational cost, and limitations in the role of decoder queries. Another line of work addresses the severe problems arising from the instability of the Hungarian matching algorithm. DN-DETR \cite{dn} introduces a novel training method that speeds up DETR training through denoising queries and attention masks. Building upon this idea, DINO \cite{dino} proposed Contrastive DeNoising Training to enhance the model's ability to suppress confusion caused by multiple anchors referring to the same object using "negative queries." Moreover, Relation DETR \cite{hou2024relationdetr} addressed the significance of object positional relationships in the detection task by introducing a position relation encoder with attention refinement. However, previous works remain computationally intensive and largely neglect the problem of real-time inference. Real-time DETR variants such as RT-DETR \cite{rtdetr} and RF-DETR \cite{robinson2026rfdetr} reduce this cost while keeping competitive accuracy.

Closest to our decoder design, DEQDet~\cite{deqdet} ties the weights of
the refinement layers in query-based detectors and solves query
refinement as a fixed point. In contrast, we unroll a single
deformable-attention decoder layer of RF-DETR for a fixed number of
iterations, initialized from pretrained weights, without a separate
initialization layer or fixed-point solver.

\subsection{Multi-Instance Interactive Segmentation}

Several existing approaches have allowed a single interaction to influence the segmentation or detection of multiple objects. DynaMITe~\cite{rana2023dynamite} encodes clicks as decoder queries and segments several instances in one pass, but each instance still requires an individual click. C3Det~\cite{lee2022c3det} extends this to interactive tiny-object detection by correlating features at class-labelled point clicks with the entire image, enabling a click to facilitate the detection of distant, unclicked objects (evaluated on Tiny-DOTA). However, its proposed clicks must be assigned a learned class by the annotator, and the propagation happens implicitly inside the network without an explicit mechanism for assessing the reliability of the detections. SAM 3~\cite{carion2026sam3} segments all instances of a concept specified by a noun phrase or box exemplars, with predictions further refined through positive and negative exemplars. Its applicability to ultra-high-resolution remote sensing imagery remains limited, as it is a general-purpose foundation model not specifically designed for tiny, densely packed objects. Furthermore, its two interaction modes operate independently: noun phrases and box exemplars provide concept-level prompts, whereas point clicks refine only the individual objects on which they are placed. Detections can also serve directly as prompts: the SAM-det baseline in RSPrompter~\cite{chen2024rsprompter} feeds a trained detector's boxes to SAM. This pipeline is fully automatic: every detection is trusted as given, so missed or false detections pass straight into the final masks, and there is no user in the loop to correct them.

Our ISRS-DETR addresses these limitations in three key aspects: (i) The user's clicks require no class label and the output is a segmented mask; (ii) Propagation is explicit rather than implicit: each same-class detection is turned into a click by transferring the user's click under four orientation hypotheses, and Dynamic Top-K decides per image how many to keep, so propagated prompts enter through the same click channel as the user's and can be inspected and corrected. (iii) It is designed for ultra-high-resolution remote sensing imagery within CrossCut's patch-wise processing framework. Our propagation mechanism can be easily plugged into any existing interactive segmentation frameworks.

\section{Method}

\begin{figure*}
    \centering    \includegraphics[width=0.9\textwidth]{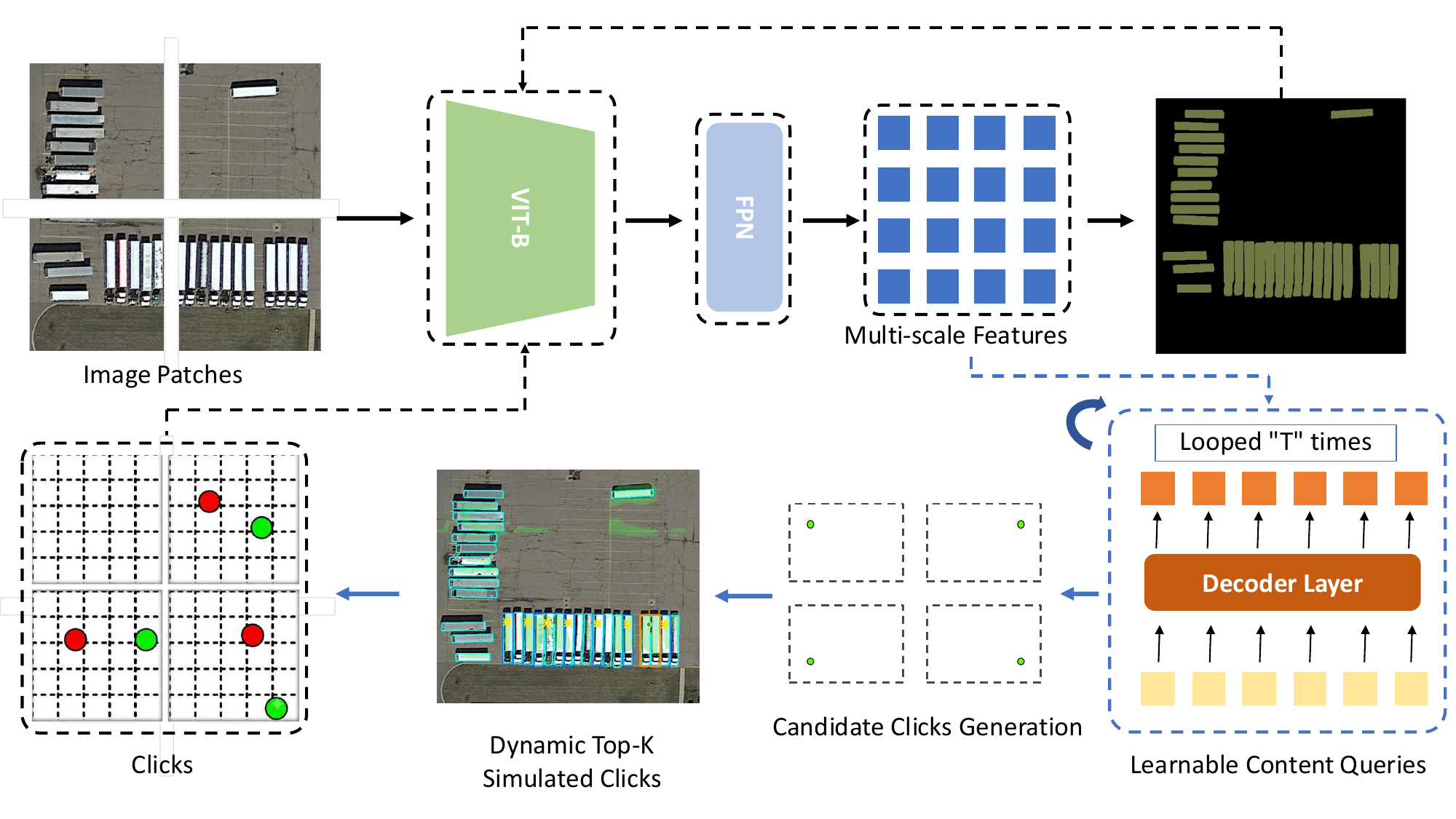}

    \caption{An overview of our \method framework: Our framework consists of two branches: (i) The segmentation branch follows CrossCut to predict patch-level masks and reconstruct the final segmentation map. (ii) The detection guide branch leverages the RF-DETR decoder with dedicated modules to propagate a single user click to all remaining similar instances of the same category.}
    \label{fig:method_overview}
\end{figure*}

\subsection{Overall Structure}

The overall architecture of our \method{} builds on CrossCut and RF-DETR. The image encoder is a plain ViT-B with a simple feature pyramid network (FPN) to produce multi-scale features similar to the original baselines. These features are then shared by two branches: (i) the segmentation branch follows CrossCut, decoding per-patch masks that are reassembled into final prediction, (ii) the detection guide branch employ the looped RF-DETR's decoder with dedicated components to propagate a single user click to every remaining appropriate instance of the same category. An overview of our proposed framework is illustrated in Figure \ref{fig:method_overview}.

\subsection{Segmentation Branch}

Following CrossCut \cite{lin2026crosscut}, the input image is resized to $H\times H$ ($H = 448\times n$), then split into an $n\times n$ grid of
non-overlapping patches $\{\mathbf{I}_1,\ldots,\mathbf{I}_{n^2}\}$. Each patch is embedded by the patch embedding
$\mathrm{E}_{img}$, and the click maps concatenated with the previous mask, $\mathbf{C}_i$, are embedded by the DistMap encoder $\mathrm{E}_{ext}$. To avoid isolating the patches from each other, a cross-patch branch derives a global Cross-Patch Prompt Embedding $\mathbf{P}$ from the click semantics and splits it into
per-patch prompts $\{\mathbf{P}_i\}_{i=1}^{n^2}$, so that every patch receives guidance even when it
contains no click. The three terms are fused by element-wise addition and processed by the shared ViT:
\begin{equation}
\begin{split}
\mathbf{F}_i &= \mathrm{E}_{img}(\mathbf{I}_i)
+ \mathrm{E}_{ext}(\mathbf{C}_i)
+ \mathbf{P}_i, \\
\{\mathbf{G}_i^{l}\}_{l=1}^{4} &= \mathrm{FPN}\big(\mathrm{ViT}(\mathbf{F}_i)\big), \\
\mathbf{S}_i &= \mathrm{Head}\big(\{\mathbf{G}_i^{l}\}\big),
\end{split}
\label{eq:seg}
\end{equation}
where $\mathbf{F}_i$ are the fused input tokens, $\{\mathbf{G}_i^{l}\}$ are the multi-scale features of patch $i$ (strides 4 to 32), and $\mathrm{Head}$ is the segmentation head of CrossCut. The patch predictions $\mathbf{S}_i$ are concatenated back into the prediction mask $\mathbf{S}$ of the whole image.

\subsection{Detection Guide Branch}
\label{detection_guide_branch_method}

\paragraph{Motivation.}
Remote sensing imagery exhibits strong inter-object correlations, where multiple instances of the same semantic category frequently appear within a single scene. Consequently, a user click on one object naturally provides valuable information about other objects of the same class. However, existing interactive frameworks neglect this property, because they treat every object of the same class as the single target, therefore allowing a user interaction to benefit only the clicked instance. Consequently, segmenting the remaining instances of that category commonly requires a fresh click, even though those instances share identical semantics and appearance. To this end, we introduce a \emph{detector-guided simulated click} strategy that propagates the user interaction to every appropriate instances by generating simulated clicks.

\paragraph{Detector Architecture.}
To identify potential same-class instances, we introduce an object detection branch based on RF-DETR~\cite{robinson2026rfdetr}. The multi-scale features ${\mathbf{G}_i^{l}}$ in Eq.~(1) already incorporate both visual and click information, providing a suitable representation for object detection without requiring a separate encoder. We therefore retain only the projector and decoder of RF-DETR. Following~\cite{robinson2026rfdetr}, the detector employs $N_q=300$ object queries per patch, with reference points initialized from the highest-scoring feature locations.

DETR-based decoders progressively refine their queries through repeated interactions with image features~\cite{detr,deformable,dino,peng2024dfineredefineregressiontask,rtdetr,robinson2026rfdetr}. Specifically, each decoder layer takes the queries from the previous layer, performs self-attention and cross-attention to the image features $\mathbf{M}$ (obtained from ${\mathbf{G}_i^{l}}$), and applies a residual update to obtain refined queries. The queries of layer $\ell$ are therefore a refined version of the queries of previous layer $\ell-1$, and all $L$ decoder layers repeat one procedure, \emph{refine the queries}, $L$ times~\cite{deformable,dino,peng2024dfineredefineregressiontask}.

Motivated by this iterative nature of this refinement process, we replace the conventional decoder stack with a single weight-shared decoder layer $\mathcal{D}_{\theta}$, comprising self-attention, deformable cross-attention to $\mathbf{M}$, and a feed-forward network. This layer is recurrently applied $T$ times, with each iteration refining the queries produced by the preceding one. We initialize $\mathcal{D}_{\theta}$ using the first layer of the pretrained RF-DETR decoder~\cite{robinson2026rfdetr}. The detector is trained with the detection loss $\mathcal{L}_{\mathrm{det}}$: Hungarian matching with a sigmoid focal classification loss, an $\ell_1$ loss and a GIoU loss, applied to every decoder output. At each iteration $r$, the resulting queries $\mathbf{q}^{(r)}$ are decoded using shared classification and bounding-box prediction heads and supervised with $\mathcal{L}_{\mathrm{det}}$. The final iteration provides the object proposals for subsequent click propagation. With $T=5$, the looped decoder preserves the depth and computational cost of the original five-layer decoder while substantially reducing its parameter count from 21.2M to 4.9M.

During training, pseudo ground-truth bounding boxes are obtained by extracting the axis-aligned bounding box of every connected component in the segmentation mask. To stabilize Hungarian matching and accelerate query convergence, we adopt the denoising-query strategy throughout training. Furthermore, during the first 20 epochs, we additionally apply the position-relation attention refinement of Relation-DETR~\cite{hou2024relationdetr}.

\paragraph{Box Filtering}
Every box, that is below a confidence threshold $\theta$ or below the minimum side length, is discarded. Moreover, after the detector define the category of the box containing the user click defines the target class $\hat{c}$, all boxes detected to be in different classes $c_j\neq\hat{c}$ are removed; if no box contains the given click, nothing is propagated in that round. Because patches are processed separately, an object that crosses a patch border is returned as two boxes that touch at the border; such pairs are merged. Finally, we employ non-maximum suppression (NMS) to eliminate duplicated detections, resulting in a filtered proposal set:
$\mathcal{B}=\{\mathbf{b}_j\}_{j=1}^{N_b}$, where $N_b$ is the number of the proposed bounding boxes.

\paragraph{Simulated Click Generation}
Let $\mathbf{u}$ be a positive user click and $\mathbf{b}_s$ the box containing it. We denote by $(a,b)\in[0,1]^2$ the click's normalized coordinates within $\mathbf{b}_s$. Since remote-sensing objects can have arbitrary orientations while the proposals are axis-aligned, we consider four possible orientations when transferring the click to another box. Specifically, for each box $\mathbf{b}_j\in\mathcal{B}$ that contains no click, we map $(a,b)$ to four candidate relative positions: $(a,b)$, $(1-b,a)$, $(1-a,1-b)$, and $(b,1-a)$, corresponding to rotations of $0^\circ$, $90^\circ$, $180^\circ$, and $270^\circ$, respectively. These positions define four candidate clicks $\mathbf{u}_j^r$, where $r\in\{0,1,2,3\}$.

\begin{table*}[t]
\centering
\caption{The NoC scores of the proposed \method{} and existing
interactive segmentation baselines on the iSAID~\cite{isaid},
WHU-Building~\cite{whu}, and NWPU VHR-10~\cite{nwpu_vhr_10} datasets. Here, "\method + Looped Decoder" is the setting that the detection decoder in \method looped five times. The best and second-best results in each column are
\textbf{boldfaced} and \underline{underlined}, respectively.}
\label{tab:noc_main}

\scriptsize
\setlength{\tabcolsep}{4.5pt}
\resizebox{\textwidth}{!}{%
\begin{tabular}{llccccccccc}
\toprule
& & \multicolumn{3}{c}{iSAID} & \multicolumn{3}{c}{WHU-Building} & \multicolumn{3}{c}{NWPU VHR-10} \\
\cmidrule(lr){3-5} \cmidrule(lr){6-8} \cmidrule(lr){9-11}
Algorithm & Backbone & 70 & 75 & 80 & 70 & 75 & 80 & 70 & 75 & 80 \\
\midrule
SimpleClick~\cite{liu2023simpleclick} & ViT-B & 14.07 & 15.05 & 16.02 & 6.24 & 7.92 & 11.80 & 3.03 & 4.33 & 5.88 \\
MFP~\cite{lee2024mfp}                             & ViT-B & 10.14 & 11.23 & 13.43 & 4.88 & 5.24 & 5.88 & \textbf{1.58} & \underline{2.10} & 3.63 \\
CrossCut~\cite{lin2026crosscut}                   & ViT-B & 9.80 & 10.85 & 12.20 & 4.76 & 5.08 & 5.68 & 1.80 & 2.11 & \underline{2.75} \\
\midrule
\method{}                                         & ViT-B & \underline{9.07} & \underline{10.25} & \underline{11.81} & \underline{1.08} & \underline{1.13} & \underline{1.21} & 1.85 & 2.11 & 2.91 \\
\method{} + Looped Decoder                        & ViT-B & \textbf{8.97} & \textbf{10.15} & \textbf{11.68} & \textbf{1.07} & \textbf{1.10} & \textbf{1.18} & \underline{1.72} & \textbf{1.98} & \textbf{2.72} \\
\bottomrule
\end{tabular}%
}
\vspace{-2mm}
\par\smallskip
\end{table*}

\begin{table*}[t]
\centering
\caption{Ablation study of the decoder design on iSAID~\cite{isaid}, WHU-Building~\cite{whu}, and NWPU VHR-10~\cite{nwpu_vhr_10}, reported as NoC ($\downarrow$) at target IoUs of 70\%, 75\%, and 80\%. The baseline is CrossCut~\cite{lin2026crosscut}. $L$ is the number of stacked decoder layers and $T$ is the number of iterations of the looped decoder. The best and second-best results in each column are \textbf{boldfaced} and \underline{underlined}, respectively.}
\label{tab:ablation_noc}
\small
\setlength{\tabcolsep}{0pt}
\begin{tabular*}{\textwidth}{@{\extracolsep{\fill}}llccccccccc@{}}
\toprule
\multirow{2}{*}[-2pt]{Method} & \multirow{2}{*}[-2pt]{Config}
 & \multicolumn{3}{c}{iSAID} & \multicolumn{3}{c}{WHU-Building} & \multicolumn{3}{c}{NWPU VHR-10} \\
\cmidrule(lr){3-5} \cmidrule(lr){6-8} \cmidrule(lr){9-11}
 & & @70 & @75 & @80 & @70 & @75 & @80 & @70 & @75 & @80 \\
\midrule
Baseline          & --        & 9.80 & 10.85 & 12.20 & 4.76 & 5.08 & 5.68 & \underline{1.80} & 2.11 & \underline{2.75} \\
+ Stacked Decoder & $L{=}5$   & 9.07 & 10.25 & 11.81 & \underline{1.08} & 1.13 & 1.21 & 1.85 & 2.11 & 2.91 \\
\midrule
\multirow{4}{*}{+ Looped Decoder}
                  & $T{=}1$   & 9.11 & 10.21 & 11.72 & \underline{1.08} & \underline{1.12} & \underline{1.20} & 1.81 & \underline{2.00} & 2.95 \\
                  & $T{=}3$   & \textbf{8.77} & \textbf{9.98} & \textbf{11.62} & \underline{1.08} & \underline{1.12} & \textbf{1.18} & 1.87 & 2.06 & 2.84 \\
                  & $T{=}5$   & \underline{8.97} & \underline{10.15} & \underline{11.68} & \textbf{1.07} & \textbf{1.10} & \textbf{1.18} & \textbf{1.72} & \textbf{1.98} & \textbf{2.72} \\
                  & $T{=}7$   & 9.06 & 10.28 & 11.83 & 1.10 & 1.15 & 1.22 & 1.94 & 2.17 & 3.06 \\
\bottomrule
\end{tabular*}
\vspace{-2mm}
\end{table*}

\paragraph{Dynamic Top-$K$ Simulated Click Selection}
Not every candidate click corresponds to a valid object instance. Directly converting all candidate clicks into simulated click would introduce noisy interacts that may degrade the segmentation performance. Furthermore, it is inappropriate to use a fixed top-k number of simulated clicks, as the number of valid same-class objects varies considerably across images.

To address these issues, we propose a "Dynamic Top-$K$ Simulated Click Selection" strategy.
First, for each candidate click $\mathbf{u}_j^r$, we obtain its feature representation $\mathbf{v}_j^r$ by bilinearly interpolating the fused feature map $\phi(\mathbf{G}_i^{2})$, where $\phi$ is the projector of the detection branch. Let $\mathbf{v}$ denote the feature vector at the original user click. Then we calculate the similarity of a candidate to the user click by using the cosine similarity:
\begin{equation}
s_j^{r} = \frac{\langle \mathbf{v}_j^{r}, \mathbf{v} \rangle}
               {\lVert \mathbf{v}_j^{r} \rVert_2 \, \lVert \mathbf{v} \rVert_2},
\end{equation}

Each box keeps only its best-matching orientation:
\begin{equation}
r_j^{*}=\arg\max_{r\in\{0,1,2,3\}} s_j^{r},\qquad
s_j=s_j^{r_j^{*}},\qquad
\mathbf{u}_j=\mathbf{u}_j^{r_j^{*}},
\label{eq:orient}
\end{equation}
so that each box contributes at most one candidate click.

The candidates whose similarity falls below $\tau$ ($=0.2$) are discarded. The remaining $m$ similarities are normalized by a softmax and sorted in ascending order, $\tilde{s}_{(1)}\le\dots\le\tilde{s}_{(m)}$. We observe that the score distribution typically exhibits a sharp transition between highly similar candidates and ambiguous ones due to the effect of softmax function, the curve of ${\tilde{S}}$ appears alike a sigmoid function. Hence, the optimal number of simulated clicks should be preserved is the minimum second-order discrete derivative of the sorted ${\tilde{S}}$ curve

\begin{equation}
\begin{aligned}
k^{*}
&=\arg\min_{1 < k < m}
\left[
(\tilde{s}_{(k+1)}-\tilde{s}_{(k)})
-(\tilde{s}_{(k)}-\tilde{s}_{(k-1)})
\right] \\
&=\arg\min_{1 < k < m}
\left(
\tilde{s}_{(k+1)}
-2\tilde{s}_{(k)}
+\tilde{s}_{(k-1)}
\right).
\end{aligned}
\end{equation}

All candidates scored at or above $\tilde{s}_{(k^{*})}$ are retained, up to the $K_{\max}$ most similar, and become the simulated clicks. When fewer than four candidates, the curve has no usable shape so all of them are retained. We set $K_{\max}=20$ for both training and inference, following the standard evaluation budget of 20 clicks in interactive segmentation (NoC@20). This ensures that click propagation does not introduce more prompts than the standard user interaction budget.

\section{Experiments}

\begin{table*}[t]
\centering
\caption{Ablation study of the decoder design on iSAID~\cite{isaid}, NWPU VHR-10~\cite{nwpu_vhr_10}, and WHU-Building~\cite{whu}, reported as IoU (\%, $\uparrow$) after 1, 5, 10, and 20 clicks. The baseline is CrossCut~\cite{lin2026crosscut}. $L$ is the number of stacked decoder layers and $T$ is the number of iterations of the looped decoder. AUC is the mean IoU over clicks 1--20. The best and second-best results for each dataset are \textbf{boldfaced} and \underline{underlined}.}
\label{tab:ablation_decoder}
\small
\setlength{\tabcolsep}{0pt}
\begin{tabular*}{\textwidth}{@{\extracolsep{\fill}}lllccccc@{}}
\toprule
\multirow{2}{*}[-2pt]{Dataset} & \multirow{2}{*}[-2pt]{Method} & \multirow{2}{*}[-2pt]{Config}
 & \multicolumn{4}{c}{mIoU (\%) at $k$ clicks} & \multirow{2}{*}[-2pt]{AUC} \\
\cmidrule(lr){4-7}
 & & & \&1 & \&5 & \&10 & \&20 & \\
\midrule
\multirow{6}{*}{iSAID}
 & Baseline          & --      & 28.01 & 57.24 & 65.45 & 69.81 & 61.29 \\
 & + Stacked Decoder & $L{=}5$ & 36.64 & 61.63 & 67.71 & 71.37 & 64.54 \\
\cmidrule(l){2-8}
 & \multirow{4}{*}{+ Looped Decoder}
                     & $T{=}1$ & 37.85 & 62.26 & 68.08 & \underline{71.78} & 65.03 \\
 &                   & $T{=}3$ & \textbf{39.85} & \textbf{63.39} & \textbf{68.78} & \textbf{72.36} & \textbf{65.93} \\
 &                   & $T{=}5$ & \underline{38.22} & \underline{62.57} & 68.05 & 71.59 & \underline{65.14} \\
 &                   & $T{=}7$ & 36.44 & 62.39 & \underline{68.09} & 71.41 & 64.83 \\
\midrule
\multirow{6}{*}{NWPU}
 & Baseline          & --      & 66.11 & 85.47 & 86.11 & 86.17 & 84.54 \\
 & + Stacked Decoder & $L{=}5$ & 77.46 & \underline{89.26} & \textbf{89.65} & \textbf{89.79} & \textbf{88.81} \\
\cmidrule(l){2-8}
 & \multirow{4}{*}{+ Looped Decoder}
                     & $T{=}1$ & 78.34 & \textbf{89.29} & \underline{89.60} & \underline{89.71} & \underline{88.80} \\
 &                   & $T{=}3$ & \textbf{78.61} & 89.06 & 89.52 & 89.65 & 88.69 \\
 &                   & $T{=}5$ & \underline{78.50} & \underline{89.26} & 89.52 & 89.62 & \textbf{88.81} \\
 &                   & $T{=}7$ & 75.98 & 88.65 & 89.01 & 89.13 & 88.08 \\
\midrule
\multirow{6}{*}{WHU}
 & Baseline          & --      & \textbf{81.73} & 89.27 & 89.33 & 89.34 & 88.87 \\
 & + Stacked Decoder & $L{=}5$ & 78.86 & \textbf{90.40} & \underline{91.41} & \underline{91.55} & \textbf{90.21} \\
\cmidrule(l){2-8}
 & \multirow{4}{*}{+ Looped Decoder}
                     & $T{=}1$ & 78.26 & \underline{90.34} & \textbf{91.42} & \textbf{91.59} & 90.16 \\
 &                   & $T{=}3$ & 78.36 & 90.28 & 91.33 & 91.49 & 90.10 \\
 &                   & $T{=}5$ & \underline{79.57} & 90.31 & 91.33 & 91.54 & \underline{90.20} \\
 &                   & $T{=}7$ & 78.10 & 90.08 & 91.26 & 91.46 & 89.98 \\
\bottomrule
\end{tabular*}
\vspace{-2mm}
\end{table*}

\paragraph{Datasets:}
\label{sec:settings}

We conduct experiments on 3 benchmark datasets iSAID \cite{isaid}, WHU-Building \cite{whu}, NWPU VHR-10 \cite{nwpu_vhr_10} to assess the model performance.

\begin{enumerate}
    \item \textbf{iSAID}: This is a large-scale aerial image dataset for instance and
    semantic segmentation, derived from DOTA~\cite{dota}. It includes 1,411 images
    for training and 458 images for validation (
    image sizes from $800\times800$ up to $4000\times13000$), and we use the
    validation images for testing. It covers 15 object categories with dense,
    multi-oriented instance masks.
    \item \textbf{WHU-Building}: It is a building extraction dataset covering Christchurch, New Zealand, containing 8,188 aerial tiles of $512\times512$
    pixels and roughly 187,000 annotated building footprints. The
    official protocol splits it into 4,736 training, 1,036 validation, and 2,416
    test tiles; as only the training and validation subsets were available to us, we
    train on the training split and report results on the validation split.
    \item \textbf{NWPU VHR-10}: It is a very-high-resolution optical remote sensing
    dataset with 800 images, of which 650 positive images contain annotated objects
    from 10 geospatial categories. Images are collected from Google Earth and pan-sharpened Vaihingen imagery, with
    sizes ranging from roughly $500\times500$ to $1100\times1100$ pixels. We use the
    instance-level mask annotations of~\citet{nwpu_vhr_10} and follow their split of
    the 650 annotated images into training and test sets.
\end{enumerate}

\paragraph{Evaluation Metrics.}
Following previous works~\cite{liu2023simpleclick, lee2024mfp, lin2026crosscut}, we adopt two performance metrics. First, we evaluate interaction efficiency using the Number of Clicks (NoC), defined as the average number of user clicks required to reach a predefined intersection-over-union (IoU) threshold. We report NoC at IoU thresholds of 70\%, 75\%, and 80\%. Second, we report the mean intersection-over-union at a fixed number of clicks (mIoU@) to evaluate segmentation quality under a predefined interaction budget.

\paragraph{Implementation Details:} All experiments are conducted on four NVIDIA A100 GPUs. Following the training protocol established in previous work~\cite{lee2024mfp, lin2026crosscut}, we train our model on all three datasets (iSAID, WHU-Building, and NWPU VHR-10), maintaining the same dataset sampling ratio of 0.4:0.35:0.25 to ensure consistency with the previous baseline. Further implementation details are provided in Section~\ref{sec:supp-impl}

\section{Results \& Analysis}

\paragraph{\method versus SOTA baselines.}
Table~\ref{tab:noc_main} compares \method with SimpleClick~\cite{liu2023simpleclick}, MFP~\cite{lee2024mfp}, and CrossCut~\cite{lin2026crosscut}. With the looped decoder, \method requires the fewest clicks in eight of the nine settings. The exception is NWPU VHR-10 at 70\% IoU, where MFP needs 1.58 clicks against our 1.72. The largest improvement is on WHU-Building: our method reaches 70\%, 75\%, and 80\% IoU with 1.07, 1.10, and 1.18 clicks, compared with 4.76, 5.08, and 5.68 for CrossCut, a reduction of up to 80\%. This gain matches our design motivation. WHU-Building contains a single category whose instances appear as dense rows of near-identical buildings, so one user click, propagated by the detection branch, guides the segmentation of that whole class. On iSAID, which spans 15 categories of small objects, \method lowers NoC from 9.80 to 8.97 at 70\% IoU and from 12.20 to 11.68 at 80\%.

\begin{table}[t]
\centering
\caption{Efficiency of the decoder variants on NWPU VHR-10~\cite{nwpu_vhr_10}, measured on an NVIDIA A100 GPU. The baseline is CrossCut~\cite{lin2026crosscut}, which has no detector. Params is the total parameter count. GFLOPs and latency are measured per interaction round.}
\label{tab:efficiency}
\small
\setlength{\tabcolsep}{0pt}
\begin{tabular*}{\columnwidth}{@{\extracolsep{\fill}}llccc@{}}
\toprule
Method & Config & Params (M) & GFLOPs & Latency (ms) \\
\midrule
Baseline          & --      & 99.2  & 8{,}873  & 1{,}064.1 \\
+ Stacked Decoder & $L{=}5$ & 137.7 & 10{,}393 & 1{,}437.5 \\
\midrule
\multirow{4}{*}{+ Looped Decoder}
                  & $T{=}1$ & 121.4 & 10{,}316 & 1{,}453.0 \\
                  & $T{=}3$ & 121.4 & 10{,}357 & 1{,}447.1 \\
                  & $T{=}5$ & 121.4 & 10{,}393 & 1{,}436.3 \\
                  & $T{=}7$ & 121.4 & 10{,}432 & 1{,}442.5 \\
\bottomrule
\end{tabular*}
\vspace{-2mm}
\end{table}

\begin{figure*}[h]
  \centering
  \includegraphics[width=\textwidth]{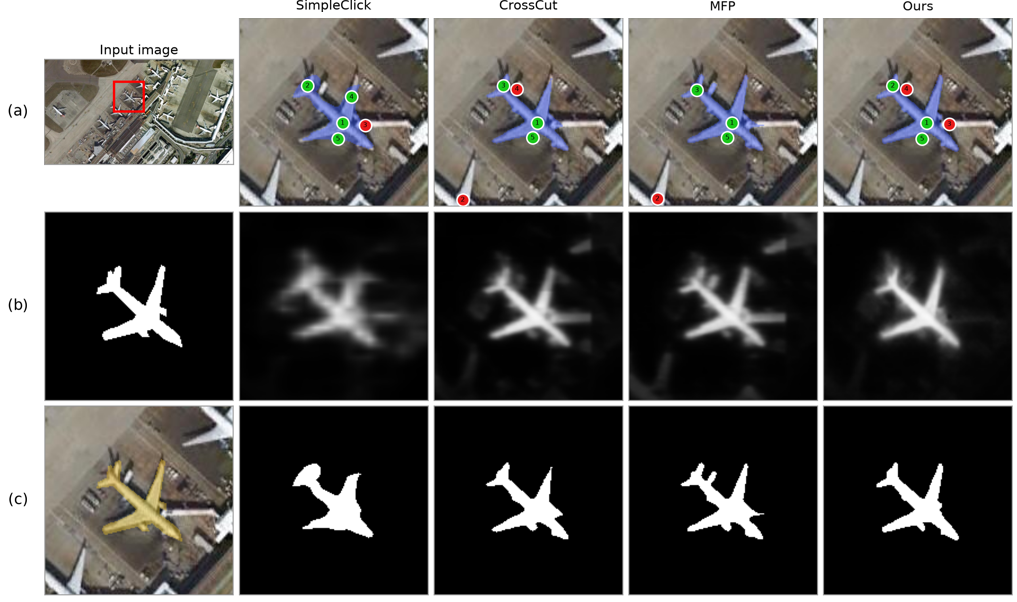}
  \caption{Qualitative comparison on NWPU VHR-10 at the same interaction
    budget of $5$ clicks. The leftmost column shows the input tile (the red box
    marks the crop used in the other columns), the ground-truth mask of the
    target object, and their overlay. For each method, row~(a) overlays the
    prediction and the simulated clicks (green: foreground, red: background,
    numbered in the order they were placed), row~(b) shows the probability map
    from the previous round, and row~(c) shows the predicted binary mask.}
  \label{fig:qualitative_nwpu}
\end{figure*}

\paragraph{Visualization Results}
Figure~\ref{fig:qualitative_nwpu} compares the four methods on a commercial aircraft from NWPU VHR-10 at an equal budget of five clicks as a hard case. Row~(c) separates the methods sharply: SimpleClick collapses the object to a fuselage blob and amputates both wings, CrossCut and MFP recover much of the wing span but leave the tips and tailplane fragmentary, and \method returns a single connected airframe with both wings and tail surfaces resolved.

\section{Ablation Study}

\paragraph{Effectiveness of Looped Detector on Segmentation Accuracy.}
Table~\ref{tab:ablation_decoder} reports mIoU after 1, 5, 10, and 20 clicks and the AUC. On iSAID, every variant of \method{} exceeds baseline (CrossCut) at every click count; the best ($T{=}3$) raises the one-click mIoU from 28.01 to 39.85 and the AUC from 61.29 to 65.93. On NWPU VHR-10, the one-click mIoU rises from 66.11 to up to 78.61 and the AUC from 84.54 to 88.81.

\paragraph{Effectiveness of Looped Detector on Segmentation Efficiency.}
Tables~\ref{tab:noc_main} and~\ref{tab:ablation_decoder} compare a stack of $L{=}5$ decoder layers against a weight-shared decoder layer recurrently applied for $T$ iterations, where $T\in\{1,3,5,7\}$ denotes the number of looping iterations. A single looped ($T{=}1$) decoder already matches the performance of the stacked 5-layer decoder. Increasing the number of iterations to $T{=}3$ yields the best results on iSAID (8.77/9.98/11.62, AUC 65.93), while $T{=}5$ performs best on WHU-Building and NWPU VHR-10. Further increasing the iterations to $T{=}7$ degrades performance across all three datasets. We therefore adopt $T{=}5$, which achieves the best or second-best results in all nine NoC settings and outperforms the stacked decoder across every NoC setting, while using only 4.9M decoder parameters compared with 21.2M.

\paragraph{Additional Computational Cost.}
Table~\ref{tab:efficiency} reports the cost per interaction round. The detection branch adds 22.2M parameters with the looped decoder and 38.5M with the stacked one, so looping saves 16.3M parameters. Each loop iteration costs about 19 GFLOPs: going from $T{=}1$ to $T{=}7$ adds 1.1\% to the total

\section{Discussion and Conclusion}

We introduce \method{}, a class-aware interactive segmentation framework for remote sensing imagery that achieves the lowest click requirements in eight of the nine evaluation settings across iSAID, WHU-Building, and NWPU VHR-10. To the best of our knowledge, we are also the first to introduce a looped decoder architecture into a DETR-based framework. Our experiments demonstrate that the looped decoder consistently outperforms a conventional five-layer stacked decoder across all NoC settings while reducing the decoder parameter count from 21.2M to 4.9M, highlighting its potential as a parameter-efficient alternative for DETR-based models.

Despite these promising results, the effectiveness of class-aware propagation depends on the number of same-class instances in a scene. Moreover, the current approach also relies on axis-aligned (horizontal) boxes, and propagates only positive clicks. Future work will explore oriented proposals, the propagation of negative and corrective clicks, learned click selection.

\newpage

\newpage

\newpage

\section*{LLM Usage Statement}
We used Claude (Anthropic), a large language model, in the following ways:
(i)~for language editing and grammar correction of author-written text; and
(ii)~for converting the \LaTeX{} source of the manuscript to the CVPR template.
All LLM-assisted code and text were reviewed and verified by the authors. All experiments, results and citations were produced and checked by the authors, who take full responsibility for the content of this paper.

{
    \small
    \bibliographystyle{ieeenat_fullname}
    \bibliography{main}

@inproceedings{lin2026crosscut,
  title     = {{CrossCut}: Cross-Patch Aware Interactive Segmentation for Remote Sensing Images},
  author    = {Lin, Zheng and Zhou, Nan and Wang, Yuhan and Zhang, Bojian},
  booktitle = {Proceedings of the AAAI Conference on Artificial Intelligence (AAAI)},
  year      = {2026}
}

@inproceedings{lee2024mfp,
  title     = {{MFP}: Making Full Use of Probability Maps for Interactive Image Segmentation},
  author    = {Lee, Chaewon and Lee, Seon-Ho and Kim, Chang-Su},
  booktitle = {Proceedings of the IEEE/CVF Conference on Computer Vision and Pattern Recognition (CVPR)},
  pages     = {4051--4059},
  year      = {2024}
}

@inproceedings{hou2024relationdetr,
  title     = {Relation {DETR}: Exploring Explicit Position Relation Prior for Object Detection},
  author    = {Hou, Xiuquan and Liu, Meiqin and Zhang, Senlin and Wei, Ping and Chen, Badong and Lan, Xuguang},
  booktitle = {Computer Vision -- ECCV 2024},
  series    = {Lecture Notes in Computer Science},
  pages     = {89--105},
  publisher = {Springer},
  year      = {2024},
  doi       = {10.1007/978-3-031-72973-7_6}
}

@inproceedings{robinson2026rfdetr,
  title     = {{RF-DETR}: Neural Architecture Search for Real-Time Detection Transformers},
  author    = {Robinson, Isaac and Robicheaux, Peter and Popov, Matvei and Ramanan, Deva and Peri, Neehar},
  booktitle = {International Conference on Learning Representations (ICLR)},
  year      = {2026}
}

@article{boykov2006graphcuts,
  title   = {Graph cuts and efficient {N-D} image segmentation},
  author  = {Boykov, Yuri and Funka-Lea, Gareth},
  journal = {International Journal of Computer Vision (IJCV)},
  year    = {2006}
}

@inproceedings{blake2004interactive,
  title     = {Interactive image segmentation using an adaptive {GMMRF} model},
  author    = {Blake, Andrew and Rother, Carsten and Brown, Matthew and Perez, Patrick and Torr, Philip},
  booktitle = {European Conference on Computer Vision (ECCV)},
  year      = {2004}
}

@article{rother2004grabcut,
  title   = {{GrabCut}: Interactive foreground extraction using iterated graph cuts},
  author  = {Rother, Carsten and Kolmogorov, Vladimir and Blake, Andrew},
  journal = {ACM Transactions on Graphics (TOG)},
  year    = {2004}
}

@inproceedings{vicente2008graph,
  title     = {Graph cut based image segmentation with connectivity priors},
  author    = {Vicente, Sara and Kolmogorov, Vladimir and Rother, Carsten},
  booktitle = {IEEE Conference on Computer Vision and Pattern Recognition (CVPR)},
  year      = {2008}
}

@inproceedings{veksler2008star,
  title     = {Star shape prior for graph-cut image segmentation},
  author    = {Veksler, Olga},
  booktitle = {European Conference on Computer Vision (ECCV)},
  year      = {2008}
}

@inproceedings{chen2022focalclick,
  title     = {{FocalClick}: Towards practical interactive image segmentation},
  author    = {Chen, Xi and Zhao, Zhiyan and Zhang, Yilei and Duan, Manni and Qi, Donglian and Zhao, Hao},
  booktitle = {IEEE Conference on Computer Vision and Pattern Recognition (CVPR)},
  year      = {2022}
}

@inproceedings{liu2023simpleclick,
  title     = {{SimpleClick}: Interactive image segmentation with simple vision transformers},
  author    = {Liu, Qin and Xu, Zhenlin and Bertasius, Gedas and Niethammer, Marc},
  booktitle = {IEEE/CVF International Conference on Computer Vision (ICCV)},
  year      = {2023}
}

@inproceedings{kirillov2023segment_sam,
  title     = {Segment anything},
  author    = {Kirillov, Alexander and Mintun, Eric and Ravi, Nikhila and Mao, Hanzi and Rolland, Chloe and Gustafson, Laura and Xiao, Tete and Whitehead, Spencer and Berg, Alexander C. and Lo, Wan-Yen and others},
  booktitle = {IEEE/CVF International Conference on Computer Vision (ICCV)},
  year      = {2023}
}

@inproceedings{xu2016deep,
  title={Deep interactive object selection},
  author={Xu, Ning and Price, Brian and Cohen, Scott and Yang, Jimei and Huang, Thomas S},
  booktitle={CVPR},
  pages={373--381},
  year={2016}
}

@misc{shan2025rossamhighqualityinteractivesegmentation,
      title={ROS-SAM: High-Quality Interactive Segmentation for Remote Sensing Moving Object}, 
      author={Zhe Shan and Yang Liu and Lei Zhou and Cheng Yan and Heng Wang and Xia Xie},
      year={2025},
      eprint={2503.12006},
      archivePrefix={arXiv},
      primaryClass={cs.CV},
      url={https://arxiv.org/abs/2503.12006}, 
}

@inproceedings{dn,
  title={Dn-detr: Accelerate detr training by introducing query denoising},
  author={Li, Feng and Zhang, Hao and Liu, Shilong and Guo, Jian and Ni, Lionel M and Zhang, Lei},
  booktitle={Proceedings of the IEEE/CVF Conference on Computer Vision and Pattern Recognition},
  pages={13619--13627},
  year={2022}
}

@article{dino,
  title={Dino: Detr with improved denoising anchor boxes for end-to-end object detection},
  author={Zhang, Hao and Li, Feng and Liu, Shilong and Zhang, Lei and Su, Hang and Zhu, Jun and Ni, Lionel M and Shum, Heung-Yeung},
  journal={arXiv preprint arXiv:2203.03605},
  year={2022}
}

@inproceedings{fpn,
  title={Feature pyramid networks for object detection},
  author={Lin, Tsung-Yi and Doll{\'a}r, Piotr and Girshick, Ross and He, Kaiming and Hariharan, Bharath and Belongie, Serge},
  booktitle={Proceedings of the IEEE conference on computer vision and pattern recognition},
  pages={2117--2125},
  year={2017}
}

@article{deformable,
  title={Deformable detr: Deformable transformers for end-to-end object detection},
  author={Zhu, Xizhou and Su, Weijie and Lu, Lewei and Li, Bin and Wang, Xiaogang and Dai, Jifeng},
  journal={arXiv preprint arXiv:2010.04159},
  year={2020}
}

@misc{detr,
      title={End-to-End Object Detection with Transformers}, 
      author={Nicolas Carion and Francisco Massa and Gabriel Synnaeve and Nicolas Usunier and Alexander Kirillov and Sergey Zagoruyko},
      year={2020},
      eprint={2005.12872},
      archivePrefix={arXiv},
      primaryClass={cs.CV},
      url={https://arxiv.org/abs/2005.12872}, 
}

@misc{rtdetr,
      title={DETRs Beat YOLOs on Real-time Object Detection}, 
      author={Yian Zhao and Wenyu Lv and Shangliang Xu and Jinman Wei and Guanzhong Wang and Qingqing Dang and Yi Liu and Jie Chen},
      year={2024},
      eprint={2304.08069},
      archivePrefix={arXiv},
      primaryClass={cs.CV},
      url={https://arxiv.org/abs/2304.08069}, 
}

@InProceedings{isaid,
  author    = {Waqas Zamir, Syed and Arora, Aditya and Gupta, Akshita and Khan, Salman and Sun, Guolei and Shahbaz Khan, Fahad and Zhu, Fan and Shao, Ling and Xia, Gui-Song and Bai, Xiang},
  title     = {iSAID: A Large-scale Dataset for Instance Segmentation in Aerial Images},
  booktitle = {Proceedings of the IEEE/CVF Conference on Computer Vision and Pattern Recognition (CVPR) Workshops},
  month     = {June},
  year      = {2019},
  pages     = {28--37}
}

@InProceedings{dota,
  author    = {Xia, Gui-Song and Bai, Xiang and Ding, Jian and Zhu, Zhen and Belongie, Serge and Luo, Jiebo and Datcu, Mihai and Pelillo, Marcello and Zhang, Liangpei},
  title     = {DOTA: A Large-Scale Dataset for Object Detection in Aerial Images},
  booktitle = {The IEEE Conference on Computer Vision and Pattern Recognition (CVPR)},
  month     = {June},
  year      = {2018}
}

@Article{whu,
  author  = {Ji, Shunping and Wei, Shiqing and Lu, Meng},
  title   = {Fully Convolutional Networks for Multisource Building Extraction
             From an Open Aerial and Satellite Imagery Data Set},
  journal = {IEEE Transactions on Geoscience and Remote Sensing},
  volume  = {57},
  number  = {1},
  pages   = {574--586},
  year    = {2019},
  doi     = {10.1109/TGRS.2018.2858817}
}

@article{nwpu_vhr_10,
  author  = {Cheng, Gong and Han, Junwei and Zhou, Peicheng and Guo, Lei},
  title   = {Multi-class geospatial object detection and geographic image classification based on collection of part detectors},
  journal = {ISPRS Journal of Photogrammetry and Remote Sensing},
  volume  = {98},
  pages   = {119--132},
  year    = {2014},
  doi     = {10.1016/j.isprsjprs.2014.10.002}
}

@misc{kingma2017adammethodstochasticoptimization,
      title={Adam: A Method for Stochastic Optimization}, 
      author={Diederik P. Kingma and Jimmy Ba},
      year={2017},
      eprint={1412.6980},
      archivePrefix={arXiv},
      primaryClass={cs.LG},
      url={https://arxiv.org/abs/1412.6980}, 
}

@misc{cordts2016cityscapesdatasetsemanticurban,
      title={The Cityscapes Dataset for Semantic Urban Scene Understanding}, 
      author={Marius Cordts and Mohamed Omran and Sebastian Ramos and Timo Rehfeld and Markus Enzweiler and Rodrigo Benenson and Uwe Franke and Stefan Roth and Bernt Schiele},
      year={2016},
      eprint={1604.01685},
      archivePrefix={arXiv},
      primaryClass={cs.CV},
      url={https://arxiv.org/abs/1604.01685}, 
}

@article{isensee2021nnunet,
  title={nnU-Net: a self-configuring method for deep learning-based biomedical image segmentation},
  author={Isensee, Fabian and Jaeger, Paul F and Kohl, Simon AA and Petersen, Jens and Maier-Hein, Klaus H},
  journal={Nature methods},
  volume={18},
  number={2},
  pages={203--211},
  year={2021},
  publisher={Nature Publishing Group}
}

@article{cheng2017remote,
  title={Remote sensing image scene classification: Benchmark and state of the art},
  author={Cheng, Gong and Han, Junwei and Lu, Xiaoqiang},
  journal={Proceedings of the IEEE},
  volume={105},
  number={10},
  pages={1865--1883},
  year={2017},
  publisher={IEEE}
}

@article{zhu2017deep,
  title={Deep learning in remote sensing: A comprehensive review and list of resources},
  author={Zhu, Xiao Xiang and Tuia, Devis and Mou, Lichao and Xia, Gui-Song and Zhang, Liangpei and Xu, Feng and Fraundorfer, Friedrich},
  journal={IEEE Geoscience and Remote Sensing Magazine},
  volume={5},
  number={4},
  pages={8--36},
  year={2017},
  publisher={IEEE}
}

@article{audebert2018beyond,
  title={Beyond RGB: Very high resolution urban remote sensing with multimodal deep networks},
  author={Audebert, Nicolas and Le Saux, Bertrand and Lef{\`e}vre, S{\'e}bastien},
  journal={ISPRS Journal of Photogrammetry and Remote Sensing},
  volume={140},
  pages={20--32},
  year={2018},
  publisher={Elsevier}
}

@misc{dutta2025aerosegharnessingsamopenvocabulary,
      title={AerOSeg: Harnessing SAM for Open-Vocabulary Segmentation in Remote Sensing Images}, 
      author={Saikat Dutta and Akhil Vasim and Siddhant Gole and Hamid Rezatofighi and Biplab Banerjee},
      year={2025},
      eprint={2504.09203},
      archivePrefix={arXiv},
      primaryClass={cs.CV},
      url={https://arxiv.org/abs/2504.09203}, 
}

@misc{lin2015microsoftcococommonobjects,
      title={Microsoft COCO: Common Objects in Context}, 
      author={Tsung-Yi Lin and Michael Maire and Serge Belongie and Lubomir Bourdev and Ross Girshick and James Hays and Pietro Perona and Deva Ramanan and C. Lawrence Zitnick and Piotr Dollár},
      year={2015},
      eprint={1405.0312},
      archivePrefix={arXiv},
      primaryClass={cs.CV},
      url={https://arxiv.org/abs/1405.0312}, 
}

@inproceedings{lee2022c3det,
title = {Interactive Multi-Class Tiny-Object Detection},
author = {Lee, Chunggi and Park, Seonwook and Song, Heon and Ryu, Jeongun and Kim, Sanghoon and Kim, Haejoon and Pereira, S{\'e}rgio and Yoo, Donggeun},
booktitle = {Proceedings of the IEEE/CVF Conference on Computer Vision and Pattern Recognition (CVPR)},
pages = {14136--14145},
year = {2022}
}

@inproceedings{rana2023dynamite,
title = {{DynaMITe}: Dynamic Query Bootstrapping for Multi-object Interactive Segmentation Transformer},
author = {Rana, Amit Kumar and Mahadevan, Sabarinath and Hermans, Alexander and Leibe, Bastian},
booktitle = {Proceedings of the IEEE/CVF International Conference on Computer Vision (ICCV)},
pages = {1043--1052},
year = {2023}
}

@inproceedings{carion2026sam3,
title = {{SAM} 3: Segment Anything with Concepts},
author = {Carion, Nicolas and Gustafson, Laura and Hu, Yuan-Ting and Debnath, Shoubhik and Hu, Ronghang and Suris, Didac and Ryali, Chaitanya and Alwala, Kalyan Vasudev and Khedr, Haitham and others},
booktitle = {International Conference on Learning Representations (ICLR)},
year = {2026},
note = {arXiv:2511.16719}
}

@article{chen2024rsprompter,
title = {{RSPrompter}: Learning to Prompt for Remote Sensing Instance Segmentation Based on Visual Foundation Model},
author = {Chen, Keyan and Liu, Chenyang and Chen, Hao and Zhang, Haotian and Li, Wenyuan and Zou, Zhengxia and Shi, Zhenwei},
journal = {IEEE Transactions on Geoscience and Remote Sensing},
volume = {62},
pages = {1--17},
year = {2024},
doi = {10.1109/TGRS.2024.3356074}
}

@misc{peng2024dfineredefineregressiontask,
      title={D-FINE: Redefine Regression Task in DETRs as Fine-grained Distribution Refinement}, 
      author={Yansong Peng and Hebei Li and Peixi Wu and Yueyi Zhang and Xiaoyan Sun and Feng Wu},
      year={2024},
      eprint={2410.13842},
      archivePrefix={arXiv},
      primaryClass={cs.CV},
      url={https://arxiv.org/abs/2410.13842}, 
}

@misc{yang2024loopedtransformersbetterlearning,
      title={Looped Transformers are Better at Learning Learning Algorithms}, 
      author={Liu Yang and Kangwook Lee and Robert Nowak and Dimitris Papailiopoulos},
      year={2024},
      eprint={2311.12424},
      archivePrefix={arXiv},
      primaryClass={cs.LG},
      url={https://arxiv.org/abs/2311.12424}, 
}

@inproceedings{deqdet,
  title     = {Deep Equilibrium Object Detection},
  author    = {Wang, Shuai and Teng, Yao and Wang, Limin},
  booktitle = {IEEE/CVF International Conference on Computer Vision (ICCV)},
  year      = {2023}
}
}

\providecommand{\nn}{--.--}
\providecommand{\tbd}[1]{\textbf{[TBD: #1]}}
\providecommand{\figdir}{Figures}

\appendix

\setcounter{section}{0}
\renewcommand{\thesection}{\Alph{section}}
\setcounter{figure}{0}\renewcommand{\thefigure}{A\arabic{figure}}
\setcounter{table}{0}\renewcommand{\thetable}{A\arabic{table}}
\setcounter{equation}{0}\renewcommand{\theequation}{A\arabic{equation}}
\setcounter{algorithm}{0}\renewcommand{\thealgorithm}{A\arabic{algorithm}}

\maketitlesupplementary

\section{Experimental Details:}
\label{sec:supp-impl}

Following existing baselines \cite{liu2023simpleclick,lin2026crosscut,lee2024mfp}, we employ the same data augmentation strategy, consisting of random resizing, cropping, horizontal flipping, rotation, and brightness adjustment.

During training, the top-300 features from the fused feature map $\{\mathbf{G}_i^{l}\}$ are selected to initialize the positional queries \cite{robinson2026rfdetr} in the detection branch. To improve the stability of Hungarian matching, we further adopt denoising queries and the attention refinement with position relation proposed in Relation-DETR \cite{hou2024relationdetr}. All models are trained for 55 epochs on the three benchmark datasets using the Adam optimizer \cite{kingma2017adammethodstochasticoptimization} with an initial learning rate of $5\times10^{-5}$.

\begin{table}[h]
\centering
\small
\caption{The hyperparameter configuration of our \method{}. We adopt the same settings as prior work: data mixture, augmentation, optimiser, schedule and epoch settings.}
\resizebox{\linewidth}{!}{%
\begin{tabular}{ll}
\toprule
\multicolumn{2}{l}{\emph{Backbone and segmentation branch}} \\
Image encoder                 & plain ViT-B + simple FPN \\
Encoder initialisation        & \texttt{cocolvis\_vit\_base} (SimpleClick) \\
Input resolution $H$          & $448\times n$ \\
Grid sizes $n$                & $2$ (train); $\{2,3,4\}$ fused at test \\
Mask decoder                  & stack of $\mathrm{Conv}_{1\times1}$ \\
Segmentation loss             & normalised focal loss ($\alpha{=}0.5$, $\gamma{=}2$) \\
\midrule
\multicolumn{2}{l}{\emph{Detection guide branch}} \\
Decoder                       & RF-DETR-XL decoder layer, $N_q=300$ \\          %
Decoder depth                 & $T=5$ looped iterations (stacked: $L=5$) \\     %
Positional queries            & Top-300 features of $\{\mathbf{G}_i^l\}$ \\
Content queries               & Learnable parameters \\
Detection loss                & cls + $\ell_1$ + GIoU (DETR) \\
Detection loss weight $\lambda$ & $0.2$ \\                                      %
\midrule
\multicolumn{2}{l}{\emph{Click propagation}} \\
Confidence threshold $\theta$ & $0.30$ \\
Min.\ side length $s_{\min}$  & $0.004$ (normalised) \\
NMS IoU threshold             & $0.6$ \\
Max detections per click    & $64$ \\
Orientation candidates/box    & 4 \\
Similarity floor $\tau$       & $0.2$ (cosine) \\                               %
Max simulated clicks $K_{\max}$ & $20$ \\                                       %
\midrule
\multicolumn{2}{l}{\emph{Optimisation}} \\
Optimiser                     & Adam ($\beta{=}(0.9,0.999)$, $\epsilon{=}10^{-8}$) \\
Initial learning rate         & $5\times10^{-5}$ \\
Weight decay                  & $0$ \\
Epochs                        & 55 \\
Batch size (per GPU / total)  & 4 / 16 \\
Max clicks per sample         & 20 \\
GPUs                          & $4\times$ NVIDIA A100 \\
\midrule
\multicolumn{2}{l}{\emph{Data}} \\
Mixture ratio                 & 0.40 : 0.35 : 0.25 \\
\bottomrule
\end{tabular}%
}

\label{tab:supp-hparams}
\end{table}

\section{Training Procedure:}

\begin{algorithm}[h]
\caption{Training procedure of \method{}}
\label{alg:training}
\textbf{Input}:

\quad $\mathbf{I}$: Input image

\quad $\mathbf{Y}$: Ground-truth class mask of $\mathbf{I}$

\quad $\mathcal{U}_0$: Initial user clicks

\quad $\mathcal{B}^{gt}$: Pseudo boxes from the ground-truth class mask $\mathbf{Y}$

\textbf{Parameter}:

\quad $N$: Number of interaction iterations

\quad $T$: Number of iterations of the looped decoder layer $\mathcal{D}_{\theta}$

\quad $\theta$: Confidence threshold for box filtering

\quad $s_{\min}$: Minimum side length for box filtering

\quad  $K_{\max}$: Maximum number of simulated clicks

\quad $\Phi$: The whole framework parameters

\textbf{Output}: total loss $\mathcal{L}$
\begin{algorithmic}[1]
\STATE $\mathcal{U}\leftarrow\mathcal{U}_0$,\;
       $\mathcal{U}_{sim}\leftarrow\emptyset$,\;
       $\mathbf{S}_{0}\leftarrow\mathbf{0}$,\;
       $\mathcal{L}\leftarrow 0$
\FOR{$t=1$ \TO $N$}
  \STATE $\mathbf{F}_i\leftarrow\mathrm{E}_{img}(\mathbf{I}_i)
          +\mathrm{E}_{ext}(\mathbf{C}_i)+\mathbf{P}_i$
          \COMMENT{$\mathbf{C}_i$ from $\mathcal{U}\cup\mathcal{U}_{sim}$, $\mathbf{S}_{t-1}$}
  \STATE $\{\mathbf{G}_i^{l}\}\leftarrow\mathrm{FPN}\big(\mathrm{ViT}(\mathbf{F}_i)\big)$,\;
         $\mathbf{S}_{t}\leftarrow\mathrm{Head}\big(\{\mathbf{G}_i^{l}\}\big)$
          \COMMENT{Eq.~(1)}
  \STATE $\{\mathcal{B}^{(r)}\}_{r=1}^{T}\leftarrow$ apply $\mathcal{D}_{\theta}$ $T$ times,\;
         $\mathcal{B}_{raw}\leftarrow\mathcal{B}^{(T)}$
  \STATE $\mathbf{u}\leftarrow\textsc{NextClick}(\mathbf{S}_{t},\mathbf{Y})$
  \STATE $\mathcal{U}\leftarrow\mathcal{U}\cup\{\mathbf{u}\}$
    \STATE $\mathcal{B}\leftarrow\textsc{Filter}
            (\mathcal{B}_{raw},\theta,s_{\min},\hat{c})$
            \COMMENT{$\hat{c}$: class of the box containing $\mathbf{u}$}
    \STATE $\mathcal{B}\leftarrow\textsc{NMS}(\mathcal{B})
            =\{\mathbf{b}_j\}_{j=1}^{N_b}$
    \STATE $\mathcal{C}\leftarrow\{\mathbf{u}_j^{r}\}$, $r\in\{0,1,2,3\}$
            \COMMENT{4 orientations per box}
    \STATE $\mathbf{v}_j^{r}\leftarrow\textsc{Interp}(\phi(\mathbf{G}_i^{2}),\mathbf{u}_j^{r})$,\; \\
           $\mathbf{v}\leftarrow\textsc{Interp}(\phi(\mathbf{G}_i^{2}),\mathbf{u})$
    \STATE $s_j\leftarrow\max_{r}\,\cos\!\big(\mathbf{v}_j^{r},\mathbf{v}\big)$
            \COMMENT{Eqs.~(2),~(3)}
    \STATE $\tilde{\mathbf{s}}\leftarrow\textsc{Sort}\big(
            \textsc{Softmax}(\{s_j \mid s_j\ge\tau\})\big)$,\; $m\leftarrow|\tilde{\mathbf{s}}|$
    \STATE $k^{*}\leftarrow\arg\min_{1<k<m}
            \big(\tilde{s}_{(k+1)}-2\tilde{s}_{(k)}+\tilde{s}_{(k-1)}\big)$
            \COMMENT{Eq.~(4)}
    \STATE $\mathcal{U}_{sim}\leftarrow$ top-$k^{*}$ candidates of $\mathcal{C}$, at most $K_{\max}$
    \STATE $\mathcal{L}\leftarrow\mathcal{L}
            +\mathcal{L}_{seg}(\mathbf{S}_{t},\mathbf{Y})
            +\lambda\sum_{r=1}^{T}\mathcal{L}_{\mathrm{det}}
            (\mathcal{B}^{(r)},\mathcal{B}^{gt})$
\ENDFOR
\STATE \textbf{return} $\mathcal{L}$
\end{algorithmic}
\label{alg:supp-training}
\end{algorithm}

Algorithm \ref{alg:supp-training} presents the
end-to-end training procedure of \method. Following existing interactive segmentation methods, $\Phi$ is trained on a simulated interaction sequence.

\begin{enumerate}
\item \textbf{Initialization} (Line 1): The set of user click $\mathcal{U}$ is
initialized, the simulated click set
$\mathcal{U}_{sim}$ is empty, and the previous predicted mask $\mathbf{S}_{0}$ is set to
zero.

\item \textbf{Iterative Clicks} (Lines 3--8): The model is finetuned on a simulated interaction sequence. In each iteration, the patch features are fused as in Eq.~(1) to predict $\mathbf{S}_{t}$, which is compared
with $\mathbf{Y}$ to place the next click on the largest error region.

\item \textbf{Candidate Click Generation} (Lines 10--16):
After, the detector generates the set of proposal boxes $\mathcal{B}_{\text{raw}}$, we filter out the low-quality boxes. In particular, all of the low-confidence boxes (confidence $<\theta$), boxes with low side lengths, and boxes whose predicted class differs from the user-selected target class $\hat{c}$ are discarded. Finally, we apply non-maximum suppression (NMS) to remove duplicate detections. Each box contributes four orientation click candidates.

\item \textbf{Simulated Click Generation} (Lines 10--15):
We rank candidate clicks by the feature similarity to the user click. Next, we determine the adaptive threshold $k^{*}$ from the sorted similarity curve, and use the top candidates as simulated clicks for the next interaction.

\end{enumerate}

Our training objective function for the  $i^{\mathrm{th}}$ click iteration is as below:

\begin{equation}
\mathcal{L}=\mathcal{L}_{\mathrm{seg}}(\mathbf{S},\mathbf{Y})
+\lambda\sum_{r=1}^{T}\mathcal{L}_{\mathrm{det}}\big(\hat{\mathcal{Y}}^{(r)},\mathcal{B}^{gt}\big),
\label{eq:loss}
\end{equation}
where $\mathcal{L}_{\mathrm{seg}}$ is the normalized focal loss between the prediction $\mathbf{S}$ and the mask $\mathbf{Y}$, $\hat{\mathcal{Y}}^{(r)}$ are the class scores and boxes predicted after the $r^{th}$ decoder iteration, and $\mathcal{B}^{gt}$ are the corresponding pseudo ground-truth boxes. $\mathcal{L}_{\mathrm{det}}$ uses Hungarian matching with a sigmoid focal classification loss, an $\ell_1$ loss and a GIoU loss, following~\cite{robinson2026rfdetr}. We set $\lambda=0.2$.

\section{Macroscopic Correlation of Object Positions}
\label{app:mc}

This section details the metric used in Fig.~\ref{fig:mc_distribution} to
quantify how strongly the objects within a single image are geometrically
related to one another. We adopt the macroscopic correlation (MC) proposed
by Hou~\cite{hou2024relationdetr} and report it for the three remote
sensing benchmarks considered in this work, with MS-COCO~\cite{lin2015microsoftcococommonobjects}.

\paragraph{Definition}

Let an image $I$ contain $N$ annotated instances, and let the $i$-th instance
be described by its axis-aligned bounding box
\begin{equation}
  \boldsymbol{b}_i = [x_i,\, y_i,\, w_i,\, h_i] \in \mathcal{R}^{4},
  \label{eq:box}
\end{equation}
where $(x_i, y_i)$ is the top-left corner and $(w_i, h_i)$ the box extent, in
pixels. Treating the objects of $I$ as the nodes of an undirected graph, we
weight the edge between nodes $i$ and $j$ by the absolute Pearson correlation
coefficient (PCC) between their box descriptors,
\begin{equation}
\begin{aligned}
\rho_{ij}
&= \left|\mathtt{Pearson}(\mathbf{b}_i,\mathbf{b}_j)\right| \\
&= \frac{\left|\sum_{k=1}^{4}(b_{ik}-\bar{b}_i)(b_{jk}-\bar{b}_j)\right|}
{\sqrt{\sum_{k=1}^{4}(b_{ik}-\bar{b}_i)^2}
 \sqrt{\sum_{k=1}^{4}(b_{jk}-\bar{b}_j)^2}},
\end{aligned}
\label{eq:pcc}
\end{equation}
with $\bar{b}_i = \tfrac{1}{4}\sum_k b_{ik}$. The macroscopic correlation of
the image is the mean edge weight of this graph, i.e. its graph intensity:
\begin{equation}
  \mathrm{MC}(I) \;=\;
  \frac{\sum_{i}\sum_{j \neq i} \rho_{ij}}{N(N-1)}.
  \label{eq:mc}
\end{equation}
Since $\rho_{ij} \in [0,1]$ and the diagonal is excluded, $\mathrm{MC}(I) \in
[0,1]$. A value of $1$ indicates that every pair of boxes is perfectly
linearly related, while $0$ indicates that no pair carries any mutual
positional information. The metric is symmetric ($\rho_{ij} = \rho_{ji}$) and
is undefined for $N < 2$.

We compute Eq.~\ref{eq:mc} for NWPU VHR-10, WHU-Building, iSAID and MS-COCO. Boxes are read directly
from the released annotations in COCO format; for iSAID, we use the official available instance annotations, for WHU-Building, we reduce the provided per-building polygons to their corresponding boxes, and for NWPU~VHR-10, under the same split used for our
experiments.

Equation~\ref{eq:pcc} quantifies the extent to which the position and spatial extent of one object can be linearly predicted from those of another. In natural image datasets such as COCO, images typically contain objects with diverse scales, aspect ratios, and spatial layouts, resulting in relatively low values of $\rho_{ij}$. In contrast, remote sensing imagery often consists of repeated instances of the same object category appearing at similar scales and following regular spatial patterns, such as rows of vehicles, arrays of storage tanks, or tiled building footprints. Consequently, object pairs within the same image exhibit substantially higher spatial correlation. As illustrated in Fig.~\ref{fig:mc_distribution}, remote sensing images consistently demonstrate stronger inter-object correlation than natural images.

\section{Additional Experimental Results:}

\subsection{Effectiveness of Dynamic Top-$K$ selection.}
\begin{table}[h]
\caption{Selection rules for simulated clicks on NWPU VHR-10~\cite{nwpu_vhr_10}, applied to identical candidate clicks with the stacked decoder ($L{=}5$). The "pseudo ground truth" (Pseudo GT) is considered as an upper bound and is excluded from the ranking. Best and second-best results are \textbf{boldfaced} and \underline{underlined}.}
\label{tab:dynamic_topk}
\small
\setlength{\tabcolsep}{0pt}
\begin{tabular*}{\columnwidth}{@{\extracolsep{\fill}}lccccc@{}}
\toprule
Rule & Kept/cls & Prec.\,$\uparrow$ & Rec.\,$\uparrow$ & Other\,$\uparrow$ & Wrong/cls\,$\downarrow$ \\
\midrule
Keep all                   & 2.59 & 0.877 & \textbf{0.907} & \textbf{0.783} & 0.32 \\
Fixed $k{=}1$              & 0.56 & \underline{0.929} & 0.206 & 0.386 & \textbf{0.04} \\
Fixed $k{=}5$              & 1.69 & 0.899 & 0.608 & 0.692 & 0.17 \\
Fixed $k{=}10$             & 2.23 & 0.885 & \underline{0.787} & \underline{0.756} & 0.26 \\
Ours (Dynamic Top-$K$)     & 1.56 & \textbf{0.942} & 0.585 & 0.501 & \underline{0.09} \\
\midrule
\textit{Pseudo GT}         & \textit{2.27} & \textit{1.000} & \textit{0.907} & \textit{0.783} & \textit{0.00} \\
\bottomrule
\end{tabular*}
\vspace{-2mm}
\end{table}

Table~\ref{tab:dynamic_topk} compares selection rules on the same candidate clicks, generated after the first user click. \textbf{Kept/cls} and \textbf{Wrong/cls} are the average numbers of retained and incorrect clicks per class. \textbf{Prec.} is the fraction of retained clicks that land on the correct clicked class, \textbf{Rec.} is the fraction of correct candidates that are retained, and \textbf{Other} is the fraction of the other same-class objects that receive a correct click.

The pseudo ground truth retains exactly the correct candidates, up to 20 clicks. Retaining all candidates achieves the highest coverage of other instances (0.783), but introduces 0.32 wrong clicks per class. This is particularly problematic because the segmentation branch cannot distinguish an incorrectly selected positive click from a user-provided click. Fixed click budgets exhibit a trade-off between selection accuracy and coverage: $k{=}1$ produces nearly error-free selections (0.04 wrong clicks) but reaches only 38.6\% of the other instances, whereas $k{=}10$ approaches the coverage of retaining all candidates, but has low precision. In contrast, Dynamic Top-$K$ consistently achieves the highest precision (0.942) while maintaining reasonable coverage of other instances.

\begin{figure*}[h]
  \centering
  \includegraphics[width=\textwidth]{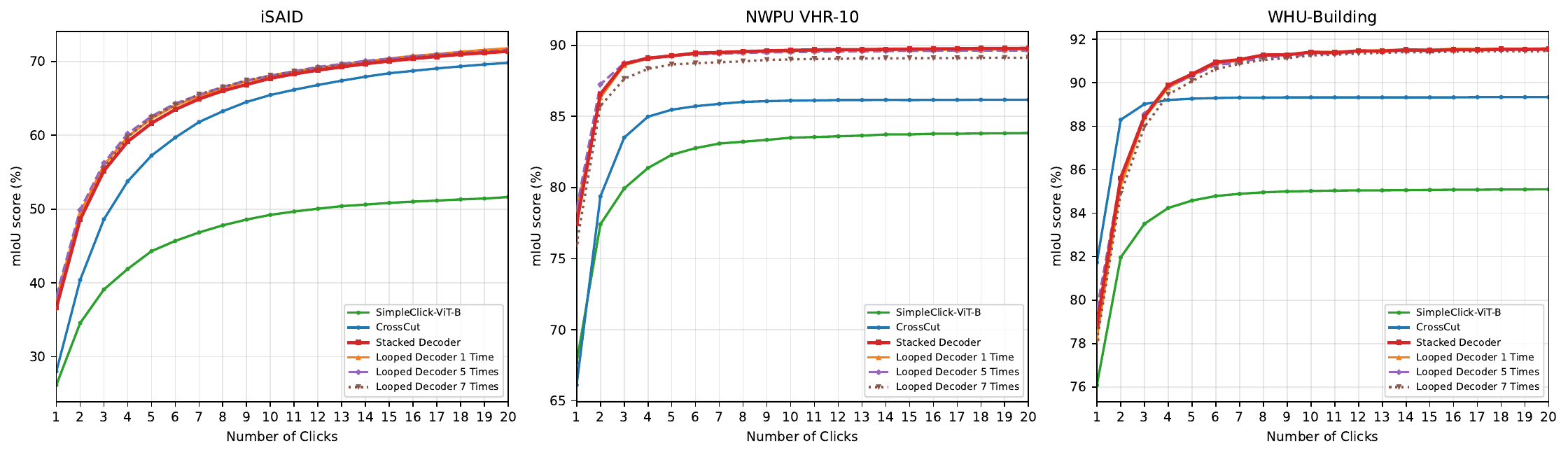}
  \caption{Mean IoU versus number of clicks on iSAID, NWPU VHR-10 and WHU-Building.
  Stacked Decoder uses five distinct decoder layers; Looped Decoder $n$ Times applies one weight-shared layer $T$ times.
  All models are trained on the same mixture of the three datasets.}
  \label{fig:miou_curves}
\end{figure*}

\subsection{Detection Quality of the Looped Decoder}
\label{sec:supp-detquality}

\begin{table}[h]
\centering
\caption{Detection quality of the detection guide branch against the pseudo ground-truth boxes on NWPU VHR-10~\cite{nwpu_vhr_10} and WHU-Building~\cite{whu}. AP$_{50}$ and Rec.$_{50}$ are the average precision and the recall at an IoU threshold of 0.5. $L$ is the number of stacked decoder layers and $T$ is the number of iterations of the looped decoder. Best and second-best results are \textbf{boldfaced} and \underline{underlined}.}
\label{tab:det_quality}
\small
\setlength{\tabcolsep}{0pt}
\begin{tabular*}{\columnwidth}{@{\extracolsep{\fill}}llcccccc@{}}
\toprule
\multirow{2}{*}[-2pt]{Decoder} & \multirow{2}{*}[-2pt]{Config}
 & \multicolumn{3}{c}{NWPU VHR-10} & \multicolumn{3}{c}{WHU-Building} \\
\cmidrule(lr){3-5} \cmidrule(lr){6-8}
 & & AP$_{50}$ & AP & Rec.$_{50}$ & AP$_{50}$ & AP & Rec.$_{50}$ \\
\midrule
Stacked & $L{=}5$ & 76.7 & 60.2 & 78.5 & 70.7 & 58.2 & 77.7 \\
\midrule
\multirow{4}{*}{Looped}
 & $T{=}1$ & 78.3 & \underline{61.0} & 81.1 & \underline{71.1} & \underline{58.6} & \underline{78.2} \\
 & $T{=}3$ & \underline{79.7} & 60.8 & \underline{82.3} & \textbf{72.5} & \textbf{59.3} & \textbf{79.7} \\
 & $T{=}5$ & \textbf{80.6} & \textbf{62.3} & \textbf{83.3} & 67.7 & 56.5 & 75.5 \\
 & $T{=}7$ & 77.0 & 60.5 & 79.6 & 69.9 & 57.8 & 77.6 \\
\bottomrule
\end{tabular*}
\vspace{-2mm}
\end{table}

Table~\ref{tab:det_quality} evaluates the detection branch itself. On NWPU VHR-10, every looped variant exceeds the stacked decoder in AP, and $T{=}5$ is the best, improving AP from 60.2 to 62.3 and AP$_{50}$ from 76.7 to 80.6. On WHU-Building, $T{=}1$ and $T{=}3$ are slightly above the stacked decoder (58.6 and 59.3 against 58.2 AP), whereas $T{=}5$ is 1.7 AP below it. Overall, the looped decoder stays within about 2 AP of the five-layer stack on both datasets while using 4.9M instead of 21.2M decoder parameters.

These results provide empirical evidence that the proposed looped transformer architecture is an appropriate alternative to the conventional stacked decoder architecture in DETR-based models, supporting the motivation discussed in Section~\ref{detection_guide_branch_method}. This makes looped decoder designs a promising direction for further investigation, particularly in exploring more effective mechanisms for iterative query refinement and parameter-efficient detection.

\subsection{Effectiveness of the Simulated Clicks}
\label{sec:supp-firstround}

\begin{table}[h]
\centering
\caption{Effectiveness of the simulated clicks generated after the first user click on iSAID~\cite{isaid}, WHU-Building~\cite{whu}, and NWPU VHR-10~\cite{nwpu_vhr_10}, with the stacked decoder ($L{=}5$) as in Table~\ref{tab:dynamic_topk}. Rows (i)--(iii) split ``No candidate kept'' by 3 reasons: (i) the class has only one instance in the image; (ii) the detector finds no other box of the same class; (iii) other boxes are found, but all of their candidate clicks fall below the similarity threshold. Wrong clicks are simulated clicks that land on another class or on the background.}
\label{tab:first_round}
\small
\setlength{\tabcolsep}{0pt}
\begin{tabular*}{\columnwidth}{@{\extracolsep{\fill}}lccc@{}}
\toprule
 & iSAID & WHU & NWPU \\
\midrule
\multicolumn{4}{l}{\emph{Outcome after the first click}} \\
No detection                                    & 0.6\%  & 0.0\%  & 1.1\%  \\
No box on the click                             & 1.7\%  & 1.0\%  & 1.7\%  \\
No candidate kept                               & 66.5\% & 50.4\% & 56.8\% \\
\hspace{1em}(i) single instance                 & 16.5\% & 8.4\%  & 38.6\% \\
\hspace{1em}(ii) no other box detected          & 22.3\% & 2.4\%  & 6.8\%  \\
\hspace{1em}(iii) all below threshold           & 27.7\% & 39.6\% & 11.4\% \\
Clicks propagated                               & 31.2\% & 48.6\% & 40.3\% \\
\midrule
Simulated clicks per sample                     & 1.27 & 3.37 & 1.56 \\
\midrule
\multicolumn{4}{l}{\emph{Location of the simulated clicks (share of clicks)}} \\
New same-class instance                         & 78.0\% & 99.5\% & 94.2\% \\
Same instance                                   & 4.0\%  & 0.2\%  & 0.0\%  \\
Tiny instance                                   & 4.0\%  & 0.0\%  & 0.0\%  \\
Other class                                     & 6.2\%  & 0.0\%  & 0.0\%  \\
Background                                      & 7.8\%  & 0.4\%  & 5.8\%  \\
\midrule
Wrong clicks                                    & 14.0\% & 0.4\%  & 5.8\%  \\
Wrong clicks per sample                         & 0.178 & 0.012 & 0.091 \\
\bottomrule
\end{tabular*}
\vspace{-2mm}
\end{table}

Table~\ref{tab:first_round} examines the effectiveness of the simulated clicks after the first user click. Simulated clicks are propagated in 31.2\%, 48.6\% and 40.3\% of the samples on iSAID, WHU-Building and NWPU VHR-10, respectively.

In the other samples no candidate is kept, for one of three reasons: (i) the class has only one instance in the image, so there is nothing to propagate to; (ii) the detector finds no other box of that class; or (iii) other boxes are found, but all of their candidate clicks fall below the similarity threshold $\tau$. The main reason differs across datasets. On NWPU VHR-10 it is reason (i), which accounts for 38.6\% of all samples. On WHU-Building it is reason (iii), with 39.6\% of the samples, so the selection itself is what withholds the clicks. On iSAID, which spans 15 categories of small objects, the three reasons are of similar size (16.5\%, 22.3\% and 27.7\%).

When clicks are propagated, they are accurate on WHU-Building and NWPU VHR-10: 99.5\% and 94.2\% of them land on a new instance of the clicked class, and only 0.4\% and 5.8\% are wrong. As iSAID is harder, 78.0\% of the simulated clicks land on a new same-class instance and 14.0\% are wrong.

\subsection{Additional Experiment on the Effectiveness of Looped Detector on Segmentation Accuracy}
\label{sec:loop}

Fig.~\ref{fig:miou_curves} compares the variants of \method{} with existing frameworks.
On all three datasets, every looped variant remains well above CrossCut and SimpleClick.
On iSAID, looping slightly improves accuracy: the AUC rises from 64.54 (stacked) to 65.03, 65.93 and 65.14 for one, three and five looped iterations, respectively, and IoU@1 rises from 36.64 to 39.85.
On NWPU and WHU, the stacked decoder and the looped variants with one to five iterations are within 0.15 AUC of each other; only seven iterations is consistently, if slightly, worse (88.08 vs.\ 88.81 on NWPU).
On WHU, all our variants start below CrossCut for the first three clicks, because a single click must cover every building in the image, and overtake it from the fourth click onward.

\subsection{Extended Qualitative Analysis of The Class-Aware Propagation}
\label{sec:supp-qualitative}

\begin{figure*}[t]

\centering
  \setlength{\tabcolsep}{1.5pt}
  \begin{tabular}{@{}ccc@{}}
    \includegraphics[width=0.328\textwidth]{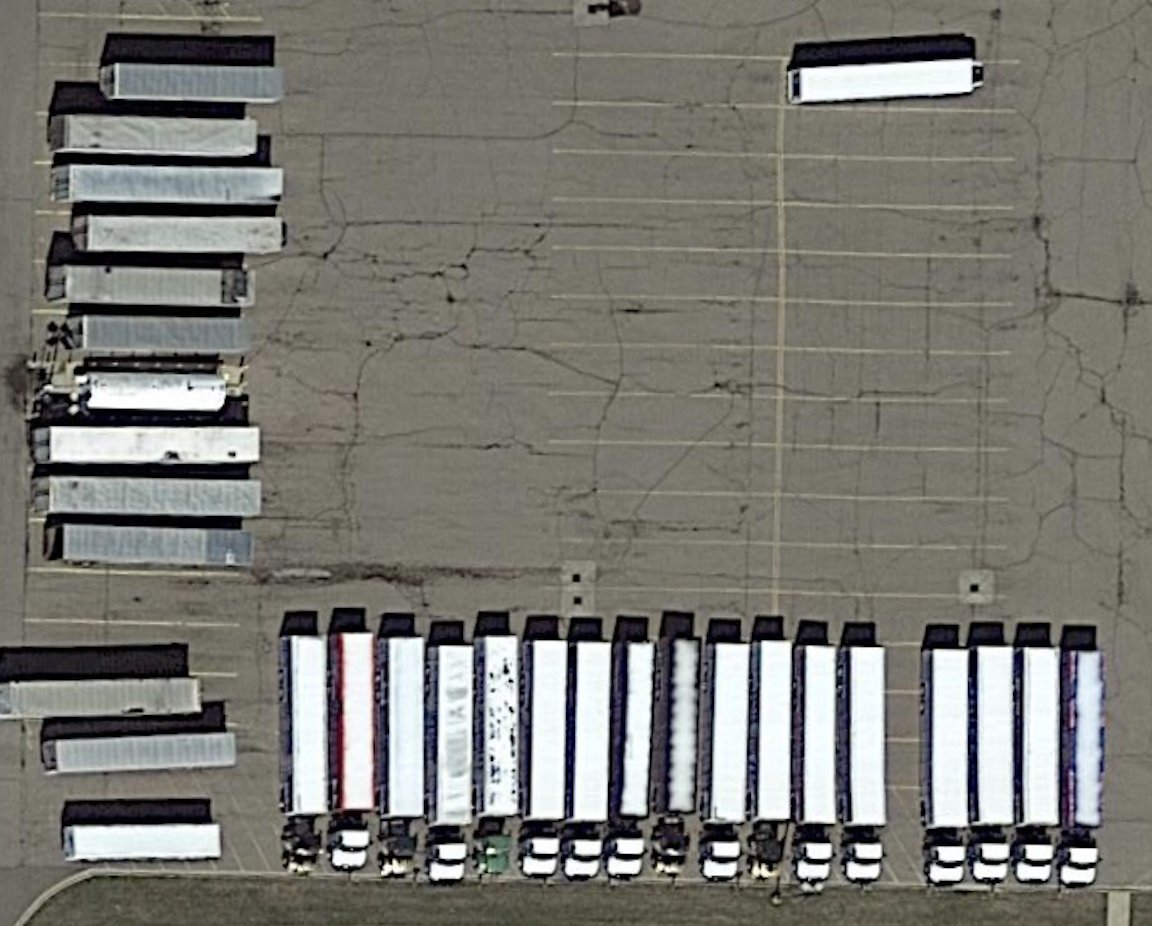} &
    \includegraphics[width=0.328\textwidth]{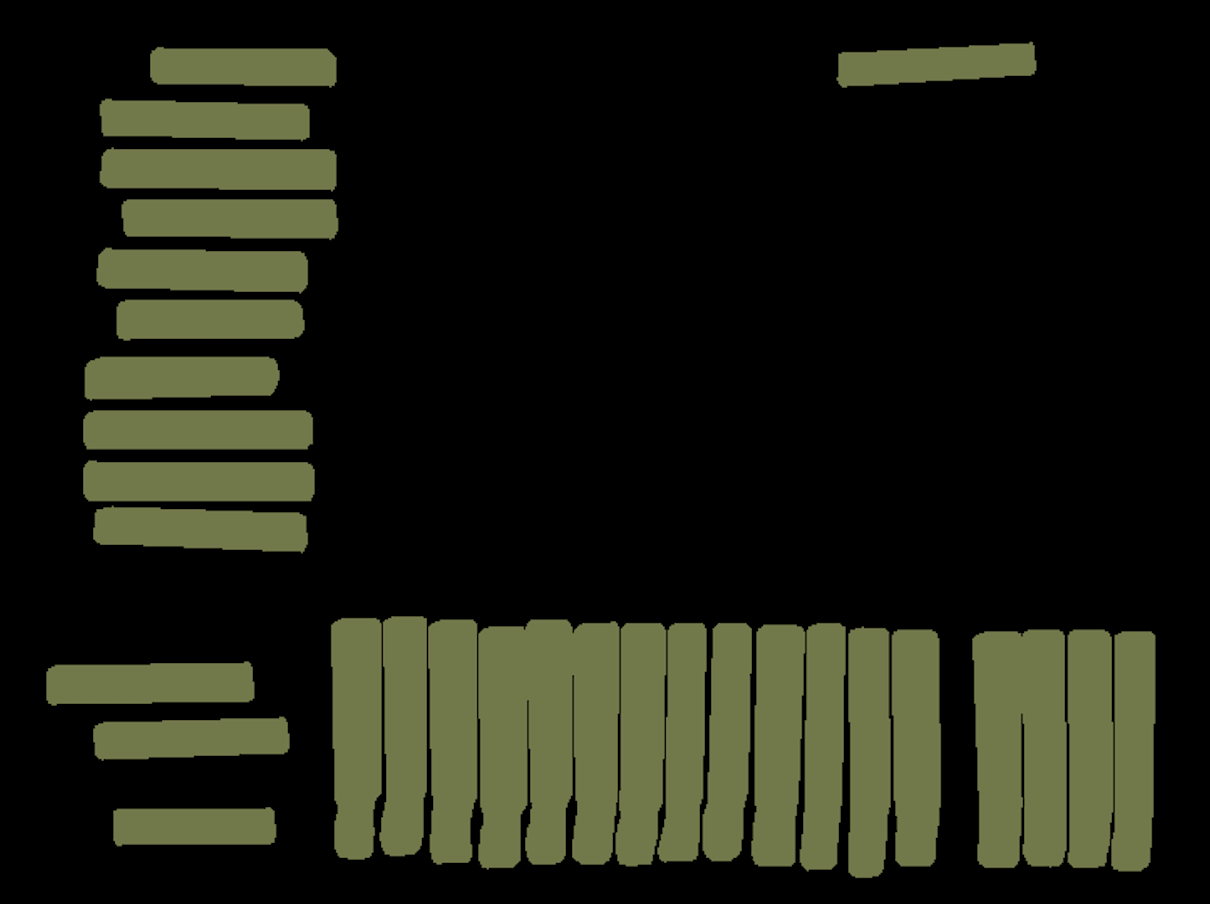} &
    \includegraphics[width=0.328\textwidth]{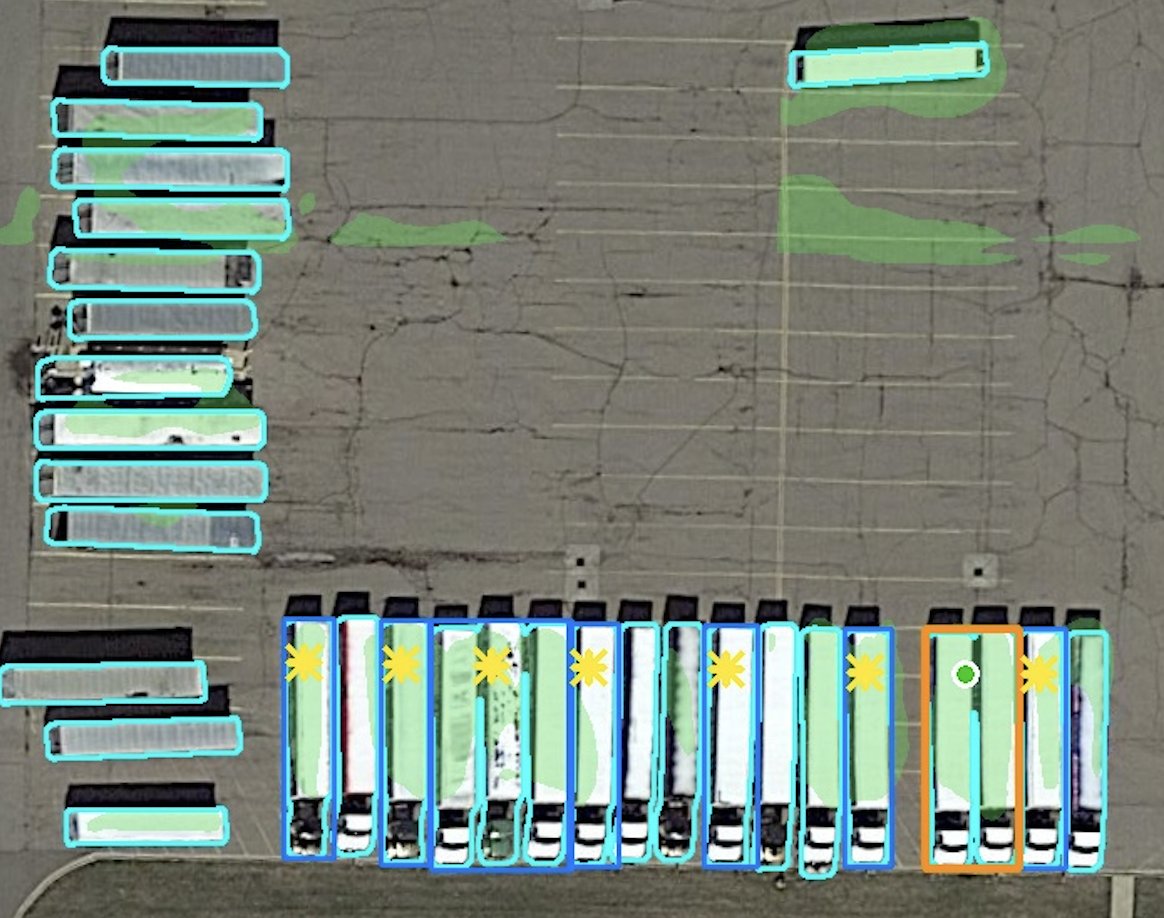} \\[2pt]
    \includegraphics[angle=90,width=0.328\textwidth]{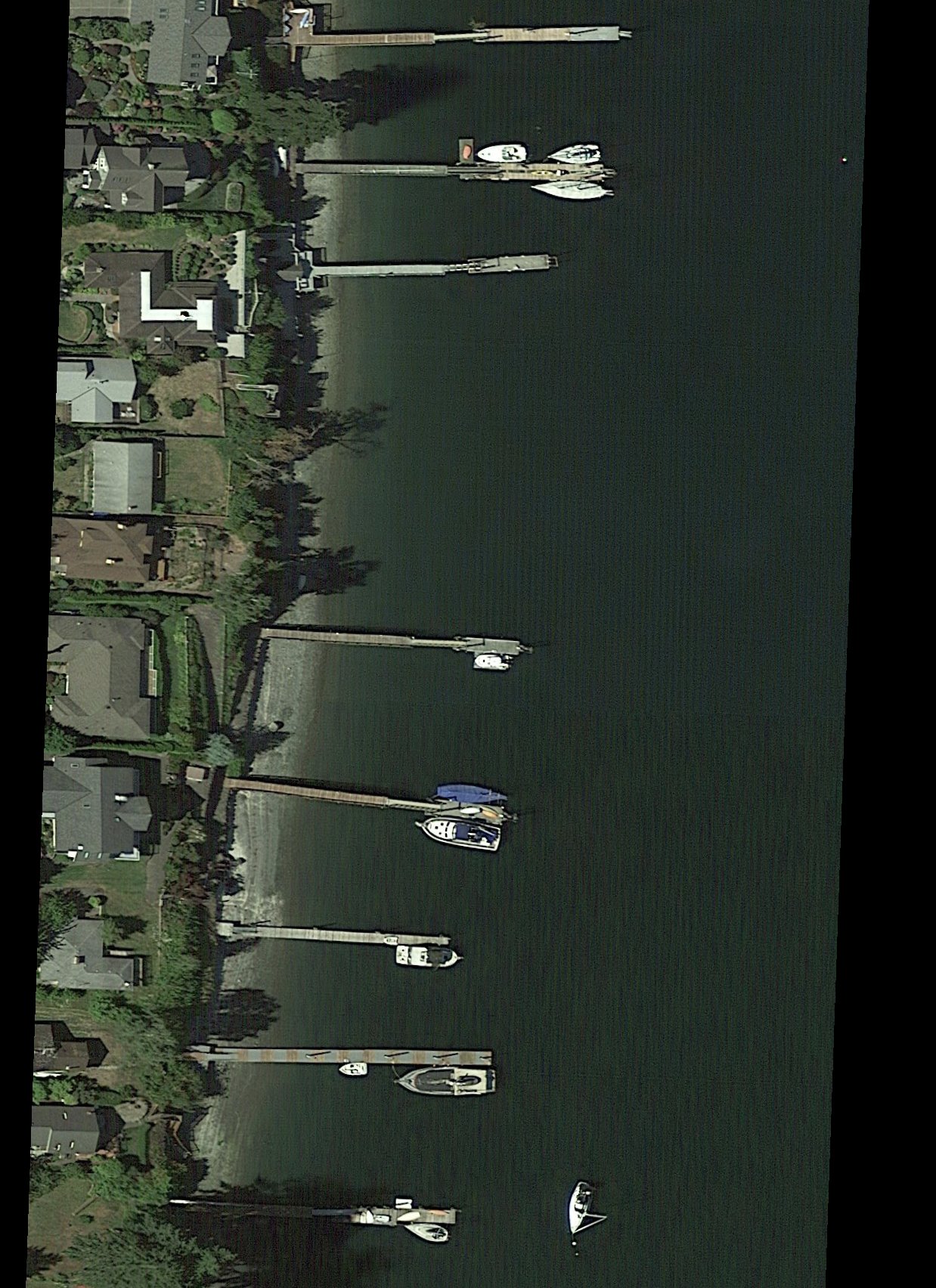} &
    \includegraphics[angle=90,width=0.328\textwidth]{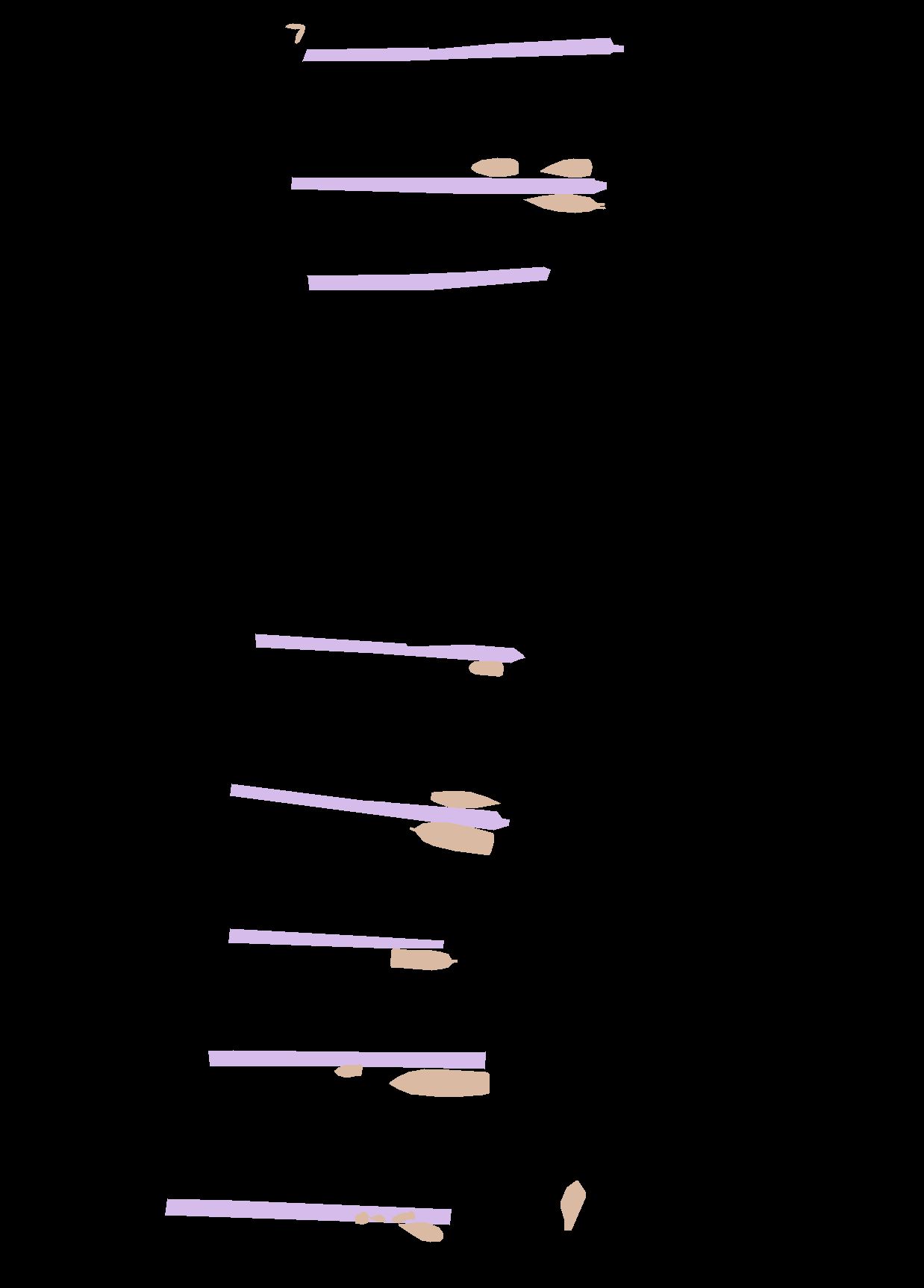} &
    \includegraphics[angle=90,width=0.328\textwidth]{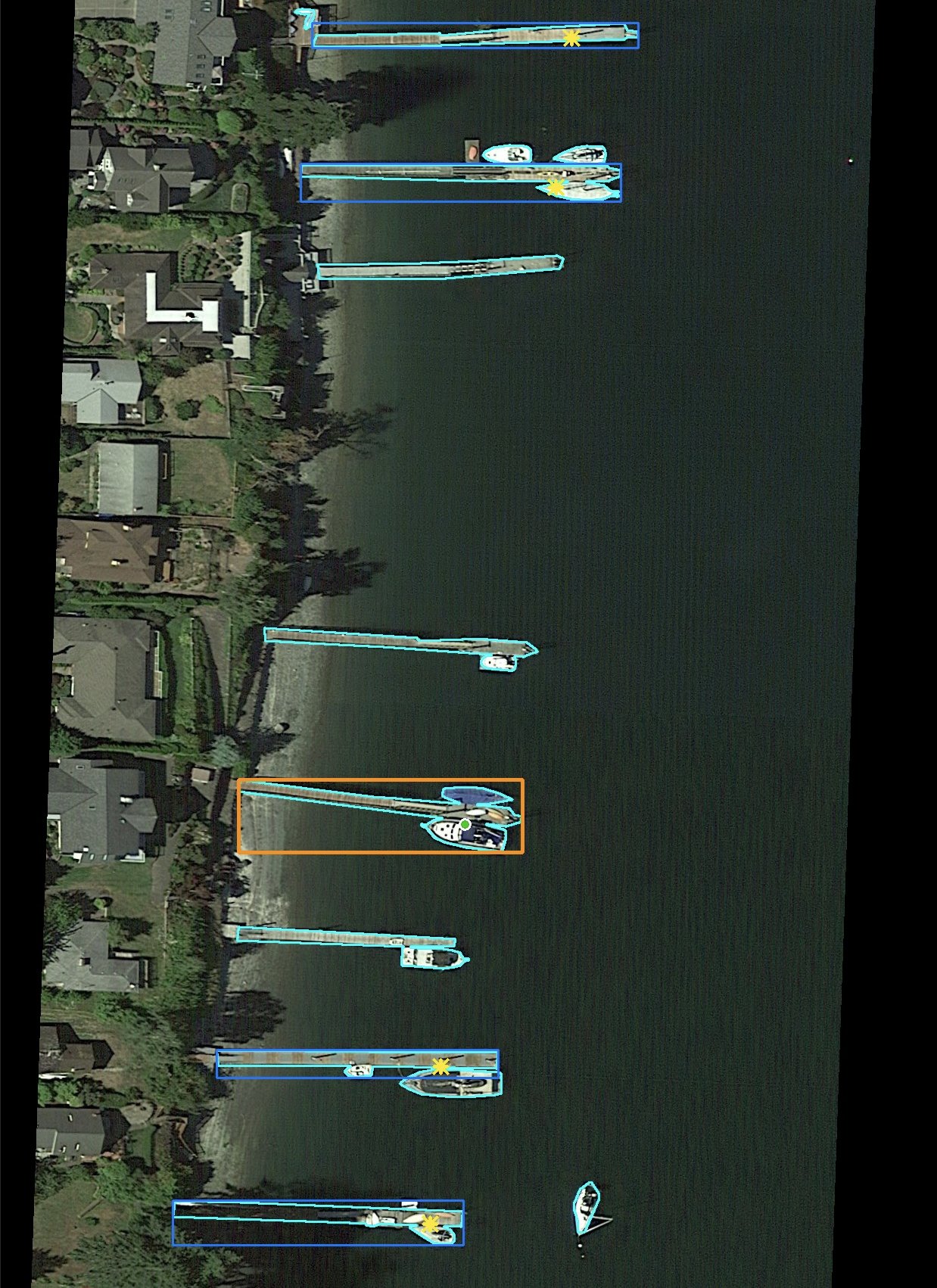} \\[2pt]
    \small (a) Input & \small (b) Ground truth & \small (c) Ours
  \end{tabular}
\caption{Class-aware propagation from a single click on iSAID. Top: a truck
depot (\emph{large vehicle}). Bottom: a residential shoreline (\emph{harbor},
rotated $90^\circ$ for layout). In (c), the orange box marks the user-clicked instance
and the green dot is the user's click; yellow markers and blue boxes indicate
co-class instances that received a simulated click from the detection branch;
cyan contours are the predicted masks and the green overlay the predicted
probability field.}
\label{fig:supp-classaware}
\end{figure*}

Figure~\ref{fig:supp-classaware} illustrates the behaviour \method{} is designed
for. In both scenes the user supplies exactly one click. The detection
branch proposes class-conditioned boxes over the entire scene, the "dynamic Top-K click strategy" selects the subset whose interpolated features are most similar to the given click, and each selected box contributes a simulated click; the segmentation
path then resolves the user click and all simulated clicks.

These two examples represent challenges commonly encountered in remote sensing imagery. The vehicle depot demonstrates a highly redundant scene containing numerous visually similar trailers. In such cases, conventional interactive segmentation commonly requires repeated user interactions for each object, causing annotation effort to grow approximately linearly with the number of instances. By exploiting the strong visual similarity among same-class objects, \method{} propagates a single user click to trailers distributed across both the densely packed row and the obliquely parked column, substantially reducing the required interactions.

In contrast, the shoreline example contains only a few instances, but each exhibits a specific geometrically complex shape. However, \method{} can still identify and propagating supervision to other semantically similar instances in the scene, demonstrating that the proposed click propagation strategy is effective across both densely populated object images and images containing geometrically complex objects.

\end{document}